\documentclass[11pt]{article}

\usepackage[preprint]{acl}

\usepackage{times}
\usepackage{latexsym}
\usepackage[T1]{fontenc}
\usepackage[utf8]{inputenc}
\usepackage{microtype}
\usepackage{inconsolata}
\usepackage{graphicx}
\usepackage{booktabs}
\usepackage{longtable}
\usepackage{multirow}
\usepackage{listings}
\lstdefinestyle{prompt}{%
  basicstyle=\scriptsize\ttfamily,%
  breaklines=true,breakindent=10pt,breakatwhitespace=false,%
  keepspaces=true,%
  frame=single,framesep=3pt,xleftmargin=3pt,xrightmargin=3pt}
\usepackage{xcolor}
\usepackage{amsmath,amssymb}
\usepackage{newunicodechar}
\newunicodechar{Δ}{$\Delta$}
\newunicodechar{β}{$\beta$}
\newunicodechar{α}{$\alpha$}
\newunicodechar{γ}{$\gamma$}
\newunicodechar{σ}{$\sigma$}
\newunicodechar{μ}{$\mu$}
\newunicodechar{≥}{$\geq$}
\newunicodechar{≤}{$\leq$}
\newunicodechar{≈}{$\approx$}
\newunicodechar{×}{$\times$}
\newunicodechar{∈}{$\in$}
\newunicodechar{∼}{$\sim$}
\newunicodechar{−}{-}
\newunicodechar{–}{--}
\newunicodechar{—}{---}

\usepackage{xspace}
\usepackage{xcolor}

\newcommand{\camerareadytext}[1]{\xspace}

\newcommand{\fref}[1]{Figure~\ref{#1}}
\newcommand{\tref}[1]{Table~\ref{#1}}

\graphicspath{{figures/}}

\newif\ifblind
\blindfalse

\newcommand{\venue}[1]{\textsc{#1}}
\newcommand{\nlpvenue}[0]{*ACL\xspace}
\newcommand{\mlvenue}[0]{\textsc{ML-general}\xspace}
\newcommand{\aivenue}[0]{\textsc{AI-broad}\xspace}
\newcommand{\myparagraph}[1]{\noindent\textbf{#1}}

\title{The Future of NLP May Not Be at NLP Conferences:\\Scholarly Migration Patterns in Natural Language Processing}

\ifblind
  \author{Anonymous ACL submission}
\else
  \author{David Jurgens \\
    School of Information, University of Michigan \\
    \texttt{jurgens@umich.edu} \\}
\fi

\begin{document}
\maketitle

\begin{abstract}

Natural Language Processing (NLP) has traditionally been published in its core disciplinary venues like ACL. However, advances in Large Language Models (LLMs) have led to a blurring of the disciplinary lines between NLP and general Machine Learning (ML), with authors regularly publishing in venues from both fields. Here, we ask whether the disciplinary center of gravity is shifting. Using NLP research published from 2010 to 2026 and studies of both established and new authors, we find that a migration is taking place. First, comparing the pre- and post-LLM eras, established authors lost 19.2~pp of share at flagship *ACL main-conference tracks while \emph{gaining} 14.8~pp in the newer Findings tracks, and general ML venues rose 8.6~pp, even when adjusting for parallel growth in the fields. Second, among newer authors who debut with at least three first-author NLP-topic papers, the share whose work appears mostly at *ACL venues fell from 84\% (2019) to 74\% (2024), while the share appearing mostly at general ML venues rose from 5\% to 21\%. Using causal inference techniques, we estimate that these general ML venues confer a significant citation premium, which influences venue selection. Together, these results point to a significant shift in where NLP research is published.

\end{abstract}

\section{Introduction}
\label{sec:intro}

The field of Natural Language Processing (NLP) has undergone a notable shift in the past decade due to the emergence of Large Language Models (LLMs). These new models offer powerful language understanding in many contexts for building applications, as well as for understanding how language is processed.
These new abilities have led to a dramatically increased growth in the field at its main conferences (ACL, NAACL, EMNLP)\footnote{We refer to this general family of venues as *ACL.}: annual *ACL output grew roughly nine-fold over the past decade, from about 975 papers in 2015 to over 8,700 in 2025. However, many of the papers introducing foundational models and techniques have not been published in NLP venues and instead appear at general Machine Learning (ML) venues; examples include GPT-3 \citep{brown2020language}, InstructGPT \citep{ouyang2022training}, Chain of Thought prompting \citep{wei2022chain}, and Chinchilla scaling laws \citep{hoffmann2022training}, which were published at NeurIPS, and LoRA \citep{hu2022lora}, FLAN \citep{wei2022finetuned}, and the ReAct agent pattern \citep{yao2023react}, which were published at ICLR.
Further, while NLP has been growing, these general ML conferences have been growing faster---over the same decade their NLP-relevant output grew roughly twenty-fold. With this growth has come a sizable number of papers on NLP topics. As a result, NLP researchers have been anecdotally said to be submitting to ICLR, NeurIPS, and ICML rather than the *ACL family of conferences, leading some to wonder whether \nlpvenue is being ``left behind.'' Here, we test this anecdote quantitatively to measure whether scholars are migrating.

To test whether NLP researchers are migrating, this paper offers the following four contributions.
First, through a large-scale analysis of 63K NLP-topic papers, drawn from 142K papers published 2010--2026 across 23 NLP, ML, and AI venues, we show that the migration is real. Among established NLP authors, publishing share at \mlvenue venues rose by 8.6~percentage points after the rise of LLMs while
their share at \nlpvenue venues fell by a comparable margin, even after
adjusting for the rapid parallel growth of both fields.
Second, using Oaxaca--Blinder decompositions, we show that this movement is due more to venue convention than to researchers changing topics.
Third, we demonstrate that new entrants into NLP research (e.g., PhD students) are increasingly likely to publish in \mlvenue venues, even when controlling for their advisor's venue preferences.
Fourth, we show that one motivation for this behavior could be due to the citation premium; using paper matching to generate counterfactual submissions, we show that a paper appearing in a \mlvenue venue is likely to receive 75--118\% more citations than it would have received if published in an \nlpvenue venue. Together, these results point to significant future changes to where the heart of NLP is and where major advancements are likely to appear.

\section{Related Work}
\label{sec:related}

Science is an evolving process where the development of new techniques has led to new fields or mergers of fields. While this case study is primarily focused on one discipline---NLP---the question of scholarly migration relates to multiple lines of work.

\myparagraph{Scholarly Incentives.}
Credit in science accrues cumulatively---also known as the Matthew effect, in which recognition flows disproportionately to already-prominent work and authors
\citep{merton-1968-matthew,price-1965-networks,azoulay-etal-2014-matthew}---so a venue that confers a citation advantage can become self-reinforcing. Closest to our setting, science-of-science work treats a researcher's field as something that moves: scientists increasingly switch topics over a career \citep{zeng-etal-2019-switch}, and interests evolve in measurable, heavy-tailed patterns \citep{jia-etal-2017-quantifying}. We read the NLP-to-ML venue shift as one consequential axis of this mobility. These works establish that venue and impact patterns are heavily author-specific and shaped by incentives, which motivates our reading of venue migration as a potential response to where the field's rewards have moved.

Differential citation rates across venues are well documented, but their
interpretation is contested. Reviews of citing behavior and of citation
indicators catalog the many non-scholarly factors at play in who gets credited for their work \citep{bornmann-daniel-2008-citation,tahamtan-etal-2016-factors,
waltman-2016-review}. These factors contribute to long-running critiques warning against reading venue-level averages as paper-level quality, such as impact factors being poor proxies for individual papers' citation counts \citep{seglen-1997-impact,garfield-2006-history} and preprint posting reshaping when and how citations accrue \citep{ginsparg-2011-arxiv,lariviere-etal-2014-arxiv}. %

\myparagraph{Bibliometrics of NLP and ML.}
A long line of work has used the ACL Anthology to study the structure and evolution of the NLP community. Multiple works have introduced new resources for studying behavior such as the Anthology Reference Corpus and the ACL Anthology Network, which turned the proceedings into a citation- and collaboration-graph resource \citep{bird-etal-2008-acl,radev-etal-2013-acl}, and the NLP Scholar dataset and explorer \citep{mohammad-2020-nlp-scholar-dataset,mohammad-2020-nlp-scholar-demo}, which focus more on the scholars. These resources have been used  to chart how the field's topical composition shifted across research epochs  \citep{anderson-etal-2012-towards}, though in the pre-LLM era. 

Building on these resources, diachronic studies have profiled productivity and impact dynamics within NLP such as the structural glass ceiling in the mentor--mentee network \citep{schluter-2018-glass}, geographic citation gaps \citep{rungta-etal-2022-geographic}, the concentration of industry labs in the field \citep{abdalla-etal-2023-elephant}, and how NLP cites and is cited by neighboring disciplines \citep{wahle-etal-2023-cite,mohammad-2020-examining}. Others trace the field's paradigm shifts and self-image directly \citep{jurgens2018measuring,pramanick-etal-2023-diachronic,michael-etal-2023-nlp,bollmann-elliott-2020-forgetting}. These studies look inward at the Anthology for how NLP authors behave within *ACL venues; we, instead, track where Anthology authors publish \emph{outside} it, and we identify a PhD-debut cohort that separates how much of the shift reflects \emph{who} is publishing from \emph{what} they work on.

\myparagraph{Research practices in NLP and ML.}
A parallel literature scrutinizes how NLP and ML conduct and report
research, often with concern about pace outrunning rigor
\citep{sculley-etal-2018-winners,lipton-steinhardt-2019-troubling}.
It documents under-powered comparisons \citep{card-etal-2020-little},
incomplete reporting of experimental results
\citep{dodge-etal-2019-show,hou-etal-2019-identification}, gaps in
reproducibility \citep{gundersen-kjensmo-2018-state}, and the value
commitments embedded in what the field chooses to study
\citep{birhane-etal-2022-values,blodgett-etal-2020-language,
rogers-etal-2021-just-think}. While not all papers in NLP and ML have such deficits, these issues contribute to the perception of what matters to reviewers in the field and what research is considered publishable.

\myparagraph{LLM-era NLP--ML convergence.}
The rise of pretrained language models (PLMs; e.g., BERT), LLMs, and other foundation models has both accelerated the growth of the field and blurred the NLP/ML boundary \citep{bommasani-etal-2021-opportunities}, against a backdrop of exponentially expanding AI literature \citep{frank-etal-2019-evolution,krenn-etal-2023-forecasting}. 
Since 2020, NLP and general ML have visibly converged around large pretrained and language models, a shift commentators have framed both methodologically and critically \citep{bender-koller-2020-climbing,bender-etal-2021-stochastic}. The institutional response is equally visible in new venues that straddle the boundary, including the launch of Transactions on Machine Learning Research \citep{tmlr-2022} and the Conference on Language Modeling \citep{colm-2024}. This convergence motivates our 2010--2026 window, which straddles the pre-PLM/LLM era up to the present; our analysis rests on a unified, cross-linked corpus built from open scholarly infrastructure \citep{lo-etal-2020-s2orc,kinney-etal-2023-semantic,priem-etal-2022-openalex}.

\section{Data}
\label{sec:data}

To model potential scholar migration, we first create a longitudinal corpus of NLP-topic papers across relevant venues with canonical cross-linked author identifiers, as described next.

\subsection{Venue taxonomy and paper coverage}
\label{sec:data_venues}

We collect data from publication venues in three categories central to the NLP$\to$ML question: \nlpvenue (e.g., ACL, EMNLP, TACL), \mlvenue (e.g., NeurIPS, ICLR, ICML), and \aivenue (AAAI, IJCAI).
*ACL itself has a diverse ecosystem, and the addition of Findings papers starting in 2020 potentially also influences the scholar migration. Therefore, we consider four tiers: \venue{NLP-Main} for all conferences' main proceedings, \venue{NLP-Findings}, \venue{NLP-Workshop/Resource}, which includes WMT and LREC,\footnote{We note that WMT is now itself a conference and LREC has always been a conference. However, both conferences have different norms from \venue{NLP-Main}, with shared tasks for WMT and a higher acceptance rate for LREC, both of which are closer to the norms of workshops.} and  \venue{NLP-Journal} for TACL and Computational Linguistics.

Paper metadata comes from a union of six sources: (i) the ACL Anthology dump,
(ii) Semantic Scholar venue queries, (iii) OpenAlex's \texttt{primary\_location}, (iv) public OpenReview submissions, (v) PMLR proceedings, and (vi) the DBLP bulk export. We deduplicate across these sources by normalizing by title within (venue, year) and using all associated identifiers for each paper (e.g., its DOI). We additionally retrieve each paper's citations, title, and abstract from Semantic Scholar and OpenAlex.
In total, we begin with 141,710 distinct papers across the three venue families, which are later classified as NLP-topic or not; \tref{tab:nlp_topic_pool} reports NLP-topic coverage within these families.
Appendix \ref{sec:appendix_data_details} contains additional details on data composition.

\subsection{Author identifier resolution}
\label{sec:data_authors}

Depending on its origin, each paper carries Semantic Scholar, OpenAlex, ACL Anthology, DBLP key, and OpenReview-id metadata where available. We cross-link author identities into a single \texttt{author\_uid} via union-find over (a)
exact-match shared IDs, (b) OpenReview profile-stated DBLP/ORCID/Google
Scholar handles, and (c) name + coauthor block uniqueness when two
records share an unambiguous canonical name. This results in 320,775 distinct
authors. Appendix Table~\ref{tab:id_coverage_nlp} reports the percent of authors with each external identifier present.

\subsection{Identifying NLP-topic Papers}
\label{sec:data_topics}

Not every paper published in \mlvenue or \aivenue is on the topic of NLP so we developed a pipeline to label papers. A \texttt{Gemma-4-26B-A4B-it} judge is prompted with the title and abstract of the paper and provided a description of possible NLP papers grounded in the call for papers from ACL venues aggregated over multiple years (details in \S\ref{sec:appendix_prompt}). The prompt was refined across multiple iterations with manual evaluation; the final version contains 2.3K tokens, with descriptions of each NLP subarea/topic, descriptions of out-of-scope topics, decision rules, and 16 examples. The model is asked to generate a terse rationale (up to 12 words), and then assign a YES/NO label of whether the paper is on an NLP topic. We note that the venue identity is masked at inference time, so the label is not collapsible to ``appeared at *ACL'' in order to prevent potentially confounding the NLP-topic label with venue in later experiments. 

To evaluate, we sample and label 402 titles and abstracts, equally balanced across \nlpvenue, \mlvenue, and \aivenue, and stratified across the Gemma 4 model's YES/NO predictions. \tref{tab:judge_eval} shows the pool-reweighted results. Performance is highest (0.96 F1) for papers in \nlpvenue and lowest for \mlvenue (0.85 F1). The residual errors are largely false positives at \mlvenue and \aivenue, on papers whose contribution is a generic ML method with an LLM as the application (e.g., quantization or decoding efficiency); genuine NLP tasks, applications, and agents at those venues are recovered at high recall.
The total number of NLP-topic papers by venue is shown in \tref{tab:nlp_topic_pool}.

  \begin{table}[t]
  \centering
  \resizebox{0.48\textwidth}{!}{
  \begin{tabular}{lrccc}
  \toprule
  Venue family & $N$ & Precision & Recall & F1 \\
  \midrule
  \textit{*}ACL    & 134 & 0.94 & 0.98 & 0.96 \\
  \mlvenue         & 134 & 0.79 & 0.92 & 0.85 \\
  \aivenue         & 134 & 0.91 & 0.88 & 0.89 \\
  \midrule
  \textit{Overall} & 402 & 0.92 & 0.96 & 0.94 \\
  \bottomrule
  \end{tabular}
  }
  \caption{Human-validated performance of the \texttt{Gemma-4-26B-A4B-it} judge (NLP = positive class), broken down by venue.}
  \label{tab:judge_eval}
  \end{table}

\begin{table}[t]
  \centering
  \small
  \begin{tabular}{lrrr}
    \toprule
    Category & Papers & NLP-topic & \% NLP-topic \\
    \midrule
    NLP        & 49,342 & 46,681 & 94.6 \\
    ML-general & 57,173 & 9,705 & 17.0 \\
    AI-broad   & 35,195 & 6,746 & 19.2 \\
    \midrule
    Total      & 141,710 & 63,132 & 44.6 \\
    \bottomrule
  \end{tabular}
  \caption{NLP-topic label coverage of the paper venue. The non-\nlpvenue NLP-topic papers are the migrating-paper pool the experiments analyze.}
  \label{tab:nlp_topic_pool}
\end{table}

\section{Are Authors Migrating?}
\label{sec:migration}

Anecdotal observations suggest that NLP researchers increasingly
publish at general ML venues since the LLM era began. Here, we empirically test this observation to assess whether there is a measurable migration, if and when it began, and whether all sub-populations of NLP authors migrate. We quantify potential migration in two ways: (1) a per-author yearly trajectory of venue-category shares, and (2) a baseline-vs-post regression of the per-author $\Delta$share broken out by author seniority. We formalize the study around three research questions (RQs): \textbf{RQ1} When did the migration begin (if ever) and is there a pre-LLM-era trend? \textbf{RQ2} How large is the shift for established NLP authors when     comparing 2015--2020 to 2021--2026? and \textbf{RQ3} Does the shift differ by author seniority?

\subsection{Experimental setup}
\label{sec:migration_setup}

\textbf{Cohort.} 
To measure migration, we first establish the cohort of authors eligible to migrate. We apply three sequential filters to our candidate pool of 320,775 authors (\S\ref{sec:data}): (1) we restrict to authors with at least one paper at a tracked-family venue (\nlpvenue, \mlvenue, or \aivenue) in the 2015--2020 baseline window (50,552 remain); (2) we further require two joint criteria---at least 3 NLP-topic papers and an NLP-topic share $\ge 50\%$ of their papers---yielding a pre-attrition pool of 5,403 authors; and (3) we keep only ``research-active'' authors, requiring at least one NLP-topic paper in 2021--2026, removing those who exited research entirely. Our final cohort consists of $N$=4,181 unique authors (Appendix Table~\ref{tab:cohort_sizes}). %

\textbf{Time windows.} We treat 2015--2020 as our baseline window; these six years
span the rise of large pretrained NLP models from early sequence-to-sequence models through GPT-3; 2021--2026 is the LLM era window. We also extend the trajectory analysis back to 2010 to bracket any pre-LLM trend in NLP-venue adherence.

\textbf{Estimator.} Our primary estimator is a single \emph{stacked}
ordinary-least-squares regression over the (author, venue category)
panel, $
  \Delta\text{share}_{ic} \;\sim\; 0 + C(\text{category})
  \;[\,+\,C(\text{category})\!:\!C(\text{stratum})\,],
$
fit with one row per author $i$ and category $c$ and cluster-robust
standard errors by author. The outcome $\Delta\text{share}_{ic}$ is the
change in author $i$'s share of papers at category $c$ from the
2015--20 baseline to the 2021--26 post window. In this two-period
collapse, the author fixed effect is period-invariant and cancels, so
each $C(\text{category})$ coefficient is the mean \emph{within-author}
pre$\to$post change in that category's venue-mix share. Stacking the
categories into one regression yields a single joint Wald test of the
null ``venue mix unchanged'' and shares the author-clustered covariance
across an author's category rows. We fit two variants of the model: (1) the venue mix share for authors regardless of their position in the author order and (2) the venue mix share only for an author's first-or-last-author papers. This latter model uses a smaller set of papers (54{,}257 vs.\ 90{,}400 author--paper records) but likely focuses the mix on papers where the author had greater influence over where the paper was published.

The two model terms answer our two estimation questions directly. \textbf{RQ2 (aggregate shift)} is read off the $C(\text{category})$ \emph{main effects}---the average within-author share change in each venue category, pooled over all established authors (Table~\ref{tab:exp1_stacked_headline}). 
\textbf{RQ3 (heterogeneity)} adds the $C(\text{category})\!:\!C(\text{stratum})$ \emph{interaction} terms, which let each category's shift vary by author stratum (Appendix Table~\ref{tab:exp1_stratum_combined}).

\textbf{Strata.} To test for heterogeneity in behavior, we use four strata: (a) career age at 2020 (junior $\leq 3$y / mid 4--8y / senior 9--15y / veteran $>$15y); (b) paper-count quartile (computed on baseline papers); (c) $h$-index quartile from Semantic Scholar / OpenAlex histories; (d) hybrid (career age $\times$ role inferred from OpenReview history when available, otherwise affiliation heuristics).

\subsection{Results: Yearly trajectory (RQ1)}
\label{sec:migration_trajectory}

\begin{figure}[t]
  \centering
  \includegraphics[width=0.47\textwidth]{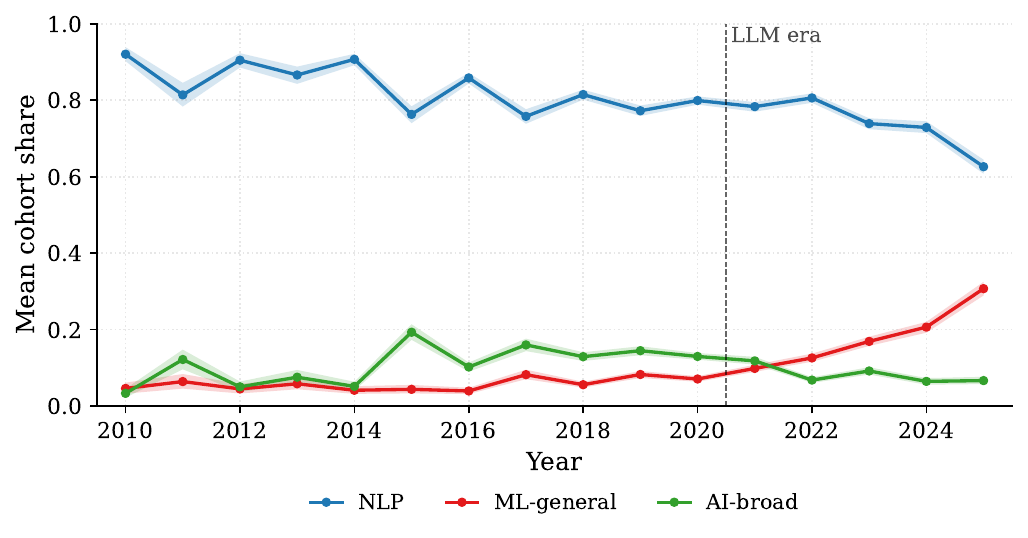}
  \caption{Per-year mean share of papers at each of the three tracked
    venue families per cohort author, renormalized to sum to one
    (topic-active cohort, $N=4{,}181$). The cohort's \nlpvenue
    share holds near 76--80\% through 2020 and then declines to
    63\% by 2025; \mlvenue rises monotonically from
    $\sim$4\% in 2015 to 31\% in 2025, overtaking \aivenue
    (which falls from 19\% to 7\%) around 2022--2023. }
  \label{fig:exp8_trajectory}
\end{figure}

Among research-active NLP researchers, a large shift is underway. As seen in \fref{fig:exp8_trajectory}, in the pre-LLM era, the \nlpvenue venues (*ACL) had  a relatively stable share of $\sim$80\% from 2015 through 2020, which then declines steadily to 63\% by 2025. Over the same window \mlvenue rises
monotonically from $\approx$4\% to 31\%, overtaking \aivenue
(which also falls from 19\% to 7\%) around 2022--2023. This shift in destinations is accelerating, pointing to a future in which \mlvenue venues may become the predominant source of NLP-topic papers.
The cutoff year of 2020 was motivated by the release of the initial LLMs. However, we performed an additional cutoff-sensitivity analysis in Appendix~\ref{sec:appendix_robustness}, which places the inflection
robustly between 2020 and 2022: sliding the pre/post boundary across candidate split years from 2017 to 2022 leaves the \nlpvenue decline and \mlvenue rise significant at each one (Table~\ref{tab:sliding_cutoff}), so the migration is not an artifact of the specific 2020 cutoff.

\subsection{Result: Per-author $\Delta$share (RQ2)}
\label{sec:migration_delta_share}

\begin{figure}[t]
  \centering
  \includegraphics[width=0.48\textwidth]{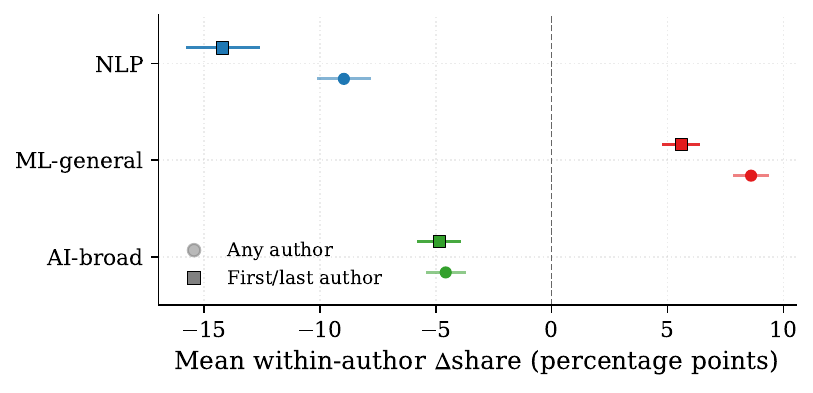}
  \caption{Mean \emph{within-author} change in renormalized venue-mix share (percentage points) from 2015--20 to 2021--26, per venue family, with 95\% cluster-robust CIs; any-authorship vs.\ first-or-last-author. Full coefficients in Appendix~\tref{tab:exp1_stacked_headline}.}
  \label{fig:exp1_stacked_headline}
\end{figure}

Both NLP and ML-general venues have seen a surge in submissions and publications; ML-general conferences are growing faster, so a potential explanation for the overall trend in \fref{fig:exp8_trajectory} is that there are simply more ML-general NLP-topic papers being produced. To rule this out, we fit the stacked regression, which isolates individual authors' behavior. Figure~\ref{fig:exp1_stacked_headline} shows the forest plot for both model specifications (coefficients in Appendix~\tref{tab:exp1_stacked_headline}), revealing sharp publication preferences by authors. The shift away from NLP venues is most pronounced when restricting to an author's first/last-authored papers, suggesting strategic behavior.

\subsection{Results: Intra-NLP Shifts}

Within \nlpvenue venues, authors have different tiers of publishing, with the ACL Organization adding a ``Findings'' venue to collect papers that were publishable but not at the level of the associated Main venue. It could be that the growth of \mlvenue is due to authors receiving a Findings decision when committing a paper to a conference (which is binding) and, instead, withdrawing to improve the paper and resubmitting it to a \mlvenue venue. Disaggregating \nlpvenue into \venue{Main}, \venue{Findings}, \venue{Workshop/Resource}, and \venue{Journal}, seen in Table~\ref{tab:exp1_tier_split}, we see this is not the case.
Instead, the disaggregation shows a stark picture for \nlpvenue: the substantial growth of Findings papers accounts for the bulk of NLP-topic papers in \nlpvenue venues. Without Findings papers, \mlvenue would account for an even larger share of papers published on NLP topics. 

\begin{table}[t]
  \centering
  \small
  \begin{tabular}{l r@{}l r@{}l}
    \toprule
    Venue tier & \multicolumn{2}{c}{Any-auth $\Delta$pp} & \multicolumn{2}{c}{First/last $\Delta$pp} \\
    \midrule
    \venue{Main} & -19.23 & $^{***}$ & -21.04 & $^{***}$ \\
    \venue{Findings} & +14.81 & $^{***}$ & +12.03 & $^{***}$ \\
    \venue{Workshop} & -4.98 & $^{***}$ & -5.29 & $^{***}$ \\
    \venue{Journal} & +0.44 & $^{*}$ & +0.12 & \\
    \venue{ML-general} & +8.60 & $^{***}$ & +5.58 & $^{***}$ \\
    \venue{AI-broad} & -4.57 & $^{***}$ & -4.84 & $^{***}$ \\
    \bottomrule
  \end{tabular}
  \caption{NLP-tier split: within-author mean $\Delta$share (2015--20 $\to$ 2021--26), in percentage points, for the topic-active cohort ($N=4,181$). The NLP category is disaggregated into Main / Findings / Workshop / Journal tiers; the denominator is the tracked venue universe. Stars use Holm--Bonferroni FWER-adjusted $p$ (within each authorship column): $^{*}p{<}0.05$, $^{**}p{<}0.01$, $^{***}p{<}0.001$ (HC1).}
  \label{tab:exp1_tier_split}
\end{table}

\subsection{Results: Author Heterogeneity (RQ3)}

Do all types of authors migrate equally? We re-fit the stacked regression separately within author strata; the two most informative---career age and $h$-index quartile---are reported in Appendix \tref{tab:exp1_stratum_combined} with Holm--Bonferroni--corrected $p$-values. The within-author \nlpvenue decline and \mlvenue gain hold in every stratum, so the migration is broad rather than confined to one group. It is also essentially uniform across career age: no career-age deviation survives correction. The one substantial exception is by citation impact---the most-cited authors are the least likely to leave \nlpvenue. Relative to the lowest $h$-index quartile, top-quartile authors show a $+8.4$pp smaller \nlpvenue decline ($p<0.001$) and shed \aivenue $6.3$pp faster ($p<0.001$), while their \mlvenue gain is unchanged.

\section{Potential Mechanism of Migration}
\label{sec:mechanism}

While authors may have multiple motivations for changing their primary venue, here we examine two potential mechanisms driving the migration: topic and author. 
As NLP and ML increasingly intersect, new opportunities to combine previously-disparate ideas may lead authors to pursue novel research directions \cite{uzzi-etal-2013-atypical}. Under the \emph{topic-led} hypothesis (H1), the topical content of NLP research has shifted toward areas (e.g., LLM scaling, RLHF, alignment) whose natural venue differs from traditional *ACL: it is not the authors who moved, but the topics they study, and the venue change follows. 
Under the \emph{author-led} hypothesis (H2), the same authors working on the same topics increasingly choose ML-general venues over NLP venues: the topical mix between venues is stable, but the venue selection is what changed.

\subsection{Experimental setup}
\label{sec:mechanism_setup}

We split each cohort venue-share change into a composition part (the cohort changed which topics it works on) and a convention part (the field changed where a given topic is published) using a Oaxaca--Blinder decomposition \citep{oaxaca1973male,blinder1973wage}, which is designed to disentangle gaps seen between two groups (here, the pre/post-LLM eras). For each tracked category $c$ and topic $t$, let $w_t^{\text{pre}}$ and $w_t^{\text{post}}$ be the fraction of cohort papers in topic $t$, and $s_{c,t}^{\text{pre}}$ and  $s_{c,t}^{\text{post}}$ be the share of topic-$t$ papers landing in category $c$ in each period. The observed cohort $\Delta$share for $c$ decomposes as
\begin{align*}
\Delta s_c \;=\;
&\underbrace{\sum_t (w_t^{\text{post}} - w_t^{\text{pre}})\,s_{c,t}^{\text{pre}}}_{\text{composition (H1)}} \\
& + \underbrace{\sum_t w_t^{\text{pre}}(s_{c,t}^{\text{post}} - s_{c,t}^{\text{pre}})}_{\text{convention (H2)}} \\
& + \underbrace{\sum_t (\Delta w_t)(\Delta s_{c,t})}_{\text{interaction}}.
\end{align*}
We report both pre-weight and post-weight references and average the two for stability. Convention measures within-topic venue substitution (supports H2 when large and signed correctly); composition measures the contribution of changes in topic mix (supports H1).

\textbf{Topic taxonomies.} We run the decomposition on two granularities of topic: (i) the OpenAlex \texttt{primary\_subfield} for each paper (\(\sim\)250 labels), which acts as a population-wide subfield assignment, and (ii) data-driven categories where we embed each title/abstract using SPECTER2 \citep{singh-etal-2023-scirepeval} and cluster with $k$-means ($k$=50). The two taxonomies are sensitivity checks on each other: they
should agree on which component dominates.

\textbf{Migrator vs.\ stayer test.} As an independent confirmation of H1 vs.\ H2, we partition the cohort into migrators ($\Delta\text{share}_{\text{NLP},i} \le -10\text{pp}$) and stayers ($|\Delta\text{share}_{\text{NLP},i}| \le 5\text{pp}$). For each author we build a baseline and a post topic-mix vector over the chosen taxonomy and compute the cosine similarity between the two. Under H1, migrators should have lower self-similarity (their topics changed); under H2, similarities should be comparable.

\subsection{Decomposition results}
\label{sec:mechanism_decomposition}

\begin{figure}[t]
  \centering
  \includegraphics[width=0.47\textwidth]{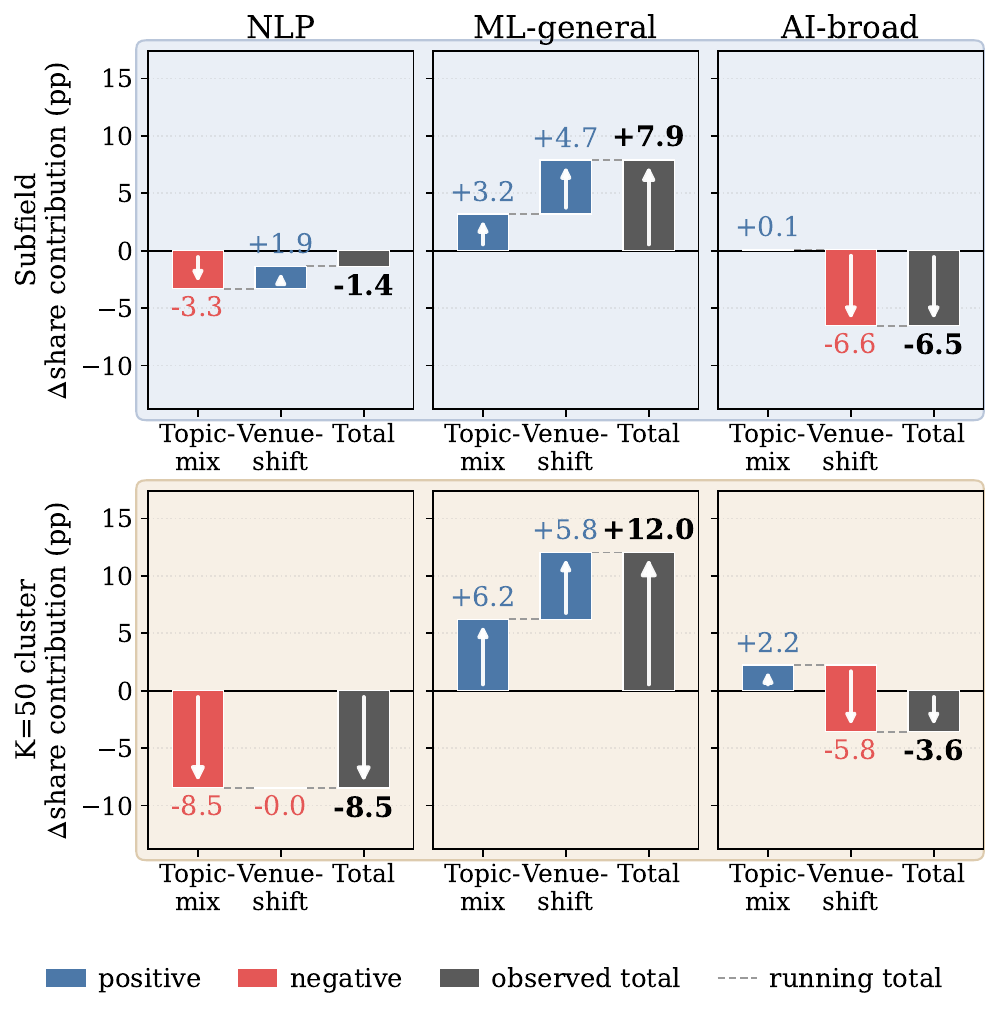}
  \caption{Results from the Oaxaca--Blinder decompositions of cohort $\Delta$share into its components for composition (topic-mix, H1; shown in red) and convention (within-topic venue, H2; shown in blue). The top uses OpenAlex subfields and the bottom uses the $k$=50 clusters.}
  \label{fig:exp15_decomposition}
\end{figure}

Note that the NLP-share movement magnitude ($-0.014$) is small because it is
paper-weighted and dominated by the heaviest NLP publishers; the within-author NLP shift in \S\ref{sec:migration} is much larger because it weights authors equally.

Using the OpenAlex subfield taxonomy, the cohort \nlpvenue share change of
-0.014 shows diverging effects: the cohort's topic mix shifted \emph{away} from NLP-friendly subfields (the larger, composition term), while within those subfields authors leaned slightly back \emph{toward} NLP venues, leaving a small net decline.
The \mlvenue gain is the topic mix moving into ML-suited subfields, but the majority is authors changing their within-topic venue convention toward ML-general. 
The \aivenue loss is almost entirely convention; authors did not stop working on AAAI/IJCAI topics; they stopped sending those topics to AAAI/IJCAI.

Both cross-family movements are convention-driven---within-topic venue substitution rather than topic change. The \mlvenue gain is majority convention ($58\%$) and the \aivenue loss is essentially all convention ($\approx$100\%): on the same topics, the cohort increasingly chose \mlvenue venues over AAAI/IJCAI. The \nlpvenue decline is the exception---it is composition-driven (the
topic mix drifted out of NLP-heavy subfields), with convention pulling
slightly the other way.

A similar result is seen with the bottom-up topic clustering. \aivenue loss continues to be convention-driven. The \nlpvenue decline is almost entirely \emph{composition}, indicating the cohort moved into content clusters that inherently under-publish at \nlpvenue venues. Together, the results suggest the cross-family \mlvenue gain and \aivenue loss are largely convention-driven, while the finer-grained taxonomy attributes more of the \nlpvenue decline to a shift in topic mix.

\subsection{Migrators vs.\ stayers}
\label{sec:mechanism_migrator_stayer}

\begin{figure}[t]
  \centering
  \includegraphics[width=\columnwidth]{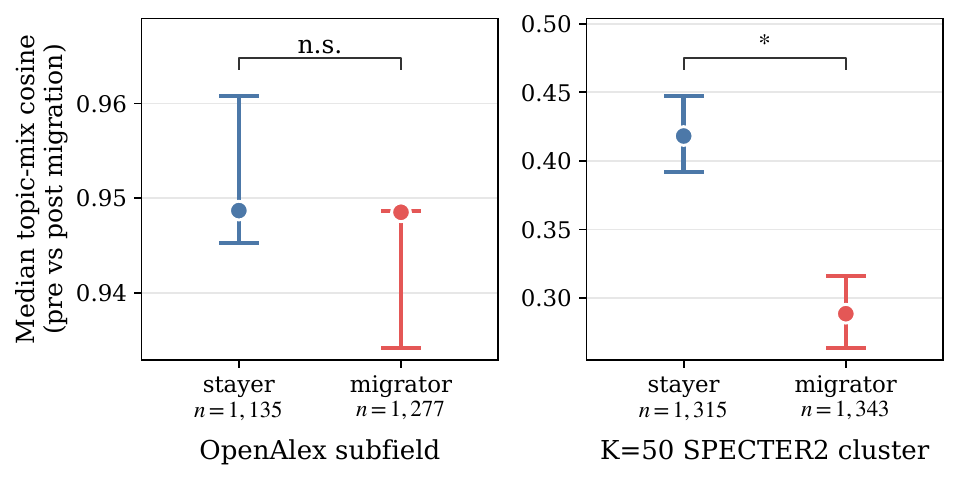}
  \caption{Median per-author cosine similarity of pre-vs-post LLM era topic-mix, stratified by migration group; error bars show bootstrapped confidence intervals and * denotes p$<$0.01. At subfield resolution, migrators and stayers are nearly identical (consistent with H2: the typical migrator kept their general subfield while switching venues); with the topically finer-grained $k{=}50$ clusters, migrators' topic mixes drift noticeably more, supporting H1, that authors are now working in topics more aligned with \mlvenue venues.}
  \label{fig:exp15_topic_stability}
\end{figure}

We find partial support for H1 (topic-led change) when examining the topic-mixes for migrators and stayers, shown in \fref{fig:exp15_topic_stability}. With the OpenAlex subfield taxonomy, both groups have a median cosine similarity of $\approx$0.95 to their earlier subfield mix, whereas with the bottom-up paper clusters the migrator--stayer gap is large and highly
significant in the H1 direction ($p=4\times10^{-11}$).
These differences reflect the granularity of the taxonomies: OpenAlex subfields are coarse (e.g., $\sim$74\% of papers fall under the single ``Artificial Intelligence'' subfield), while the bottom-up clusters capture finer, more track-like groupings. At that finer level, migrators' topic mixes drifted more than stayers', consistent with a topic-level component to the migration.
Appendix Figure \ref{fig:exp15_topic_time_series} shows how the most common clusters' \nlpvenue shares changed over time, with declines for clusters covering LLMs, reasoning, and LLM efficiency.

\section{Where do new NLP Authors Debut?}
\label{sec:phd_debut}

The previous study was focused on established NLP researchers. However, new researchers regularly enter the community, and their venue preferences also drive the movement of the field. As a complementary analysis, we ask whether these new entrants to the field show the same pattern: when a researcher publishes their first three NLP-topic papers, where are these likely to appear, and has that destination changed over 2019--2024?

\subsection{Experimental Setup}

\paragraph{Identifying New NLP Entrants.}
Identifying new entrants is itself a measurement choice. We report two cohorts and hold the statistical design fixed across them.
(1) The \emph{publication-record} cohort ($N$=3,568) includes every author with at least three first-author NLP-topic papers at a tracked venue whose first such paper falls in the study window; the author's entry year is the year of that first NLP-topic first-author paper.
(2) The \emph{declared-PhD} cohort ($N$=1,228) includes researchers whose OpenReview, ORCID, or DBLP-thesis profile parses to a doctoral start year in the window and who likewise have at least three first-author NLP-topic papers; entry year is the declared PhD start.
Requiring at least three first-author NLP-topic papers in both cohorts ensures we study where committed NLP newcomers \emph{choose} to publish, not whether they research in NLP, and it fixes the number of analyzed papers at three per author. Full details of the cohort construction process are in Appendix \ref{sec:appendix_phd_cohort}.

The two cohorts each have their own trade-offs. The declared-PhD cohort has relatively precise entry markers, access to richer metadata through OpenReview, and, because these are PhD students, likely reflects the individuals who will make up the future composition of the field. However, the declared-PhD cohort inherits the coverage biases of those platforms; OpenReview was primarily an \mlvenue platform before *ACL venues began switching to it in 2021, while ORCID is more common among European researchers, which both introduce selection-effect biases.
In contrast, the publication-record cohort avoids platform biases. However, it includes authors who may not continue on in NLP (e.g., undergraduate and masters students) and therefore may be less likely to generalize.

\paragraph{Controlling for Advisor Influence.}
Advisors likely shape the path of early-career researchers and therefore influence which venue a student submits to. Within our data, a student's advisor is recovered only from the advisor field of an OpenReview profile, which we resolve to that advisor's publication record. To control for the advisor's venue preference, we include a control variable for the advisor's share of NLP-topic papers in the five years before the
student's entry (together with an indicator for whether any advisor record was found). A computable advisor record is available for 374 of the 1,228 declared-PhD students (31\%).

\paragraph{Unit of analysis and dependent variable.}
We fit separate student-level logit models (one per cohort), each row a researcher, sharing the \emph{same} dependent variable: $Y_i=\mathbf{1}$ if a majority of researcher $i$'s first three NLP-topic first-author papers are at \nlpvenue venues.
Both models use the entry cohort year as the regressor of interest. Because every author contributes exactly three papers, no paper-count control is needed. The declared-PhD model additionally includes the advisor effects. Because the two models share the dependent variable, the cohort-year regressor, and the estimator, their cohort-year coefficients are directly comparable.

\subsection{Results}
\label{sec:phd_setup_platform}

\begin{table}[t]
  \centering
  \small
  \begin{tabular}{l r@{}l r@{}l}
    \toprule
    Predictor & \multicolumn{2}{c}{Publication-record} & \multicolumn{2}{c}{Declared-PhD} \\
    \midrule
    Cohort year & -0.166 & $^{***}$\,\tiny{(0.029)} & -0.119 & $^{**}$\,\tiny{(0.046)} \\
    Advisor NLP rate & \multicolumn{2}{c}{---} & +1.383 & $^{***}$\,\tiny{(0.315)} \\
    Advisor known & \multicolumn{2}{c}{---} & -0.699 & $^{***}$\,\tiny{(0.190)} \\
    \midrule
    Students ($N$) & \multicolumn{2}{c}{3,568} & \multicolumn{2}{c}{1,228} \\
    \bottomrule
  \end{tabular}
  \caption{Student-level logit models for whether the majority of a student's first three NLP-topic first-author papers are at \nlpvenue venues.
  Both models show the same significant decline in NLP-venue debut across cohort years, and in the declared-PhD model that decline is essentially unchanged when the advisor controls are dropped ($-0.125$). $^{*}p{<}0.05$, $^{**}p{<}0.01$, $^{***}p{<}0.001$.}
  \label{tab:exp2_unified}
\end{table}

The results from both cohort definitions, seen in \tref{tab:exp2_unified}, indicate a consistent decline in the odds that the majority of the new entrant's first three NLP-topic papers appear in \nlpvenue venues. In the publication-record cohort the share debuting mostly at \nlpvenue falls from 84\% (2019) to 74\% (2024) while the \mlvenue share rises from 5\% to 21\%. Looking at the declared-PhD cohort, we see that this effect survives the influence of the advisor's prior \nlpvenue paper distribution; even though having an established \nlpvenue-publishing advisor makes a student more likely to publish in that venue, there is still a general trend toward \mlvenue. The advisor's NLP rate is a strong positive predictor
($\beta = +1.38$, $p < 10^{-5}$): net of the cohort-year decline,
students advised by predominantly-NLP faculty are markedly more likely
to debut at NLP venues, indicating that the generational shift operates
on top of---not merely through---advisor topic inheritance.

\section{The Citation Premium}
\label{sec:citation_premium}

If multiple venues are a topical fit, why might a scholar move? Prior work suggests that authors seek out status and reputation rewards in choosing where to publish, with one common driver being the impact factor of the venue \citep{tenopir2016motivates,salinas2015should}; indeed, the most-cited authors in an area migrate the least topically \citep{petersen-etal-2014-reputation}. One natural, observable reward to model is the number of citations a paper receives. If publishing at an \mlvenue venue produces meaningfully more citations than publishing the same paper at an \nlpvenue venue, the migration could be partly explained by a rational author response to differing citation rewards. Testing this experimentally is nearly impossible as two identical papers would need to be published in different venues to see which attracts more citations; however, here, we draw from causal inference techniques to match similar papers across venues in order to estimate the citation premium by venue category, if any.

\subsection{Experimental Setup}
We fit two versions of a regression model where the outcome is $\log(1+\text{cites})$ using Semantic Scholar counts. To control for paper topic, we include a fixed effect for its cluster membership from the $k$=50 SPECTER2 topic-cluster embedding; we also account for overall growth in the field with publication-year fixed effects.
We restrict our analysis to \nlpvenue and \mlvenue first- or last-authored papers, and compare their citations \emph{within cluster and year}. The two models differ only in how matching is handled:
(1) \emph{All papers}: we match each \mlvenue paper to its three     same-year nearest \nlpvenue neighbors corpus-wide (median cosine     similarity $0.92$), pool the matched papers (controls down-weighted by     $1/3$), and regress on an \mlvenue indicator with year fixed     effects. The nearest-neighbor match serves as a content control, i.e., a counterfactual of how the \mlvenue paper would have done if published in an \nlpvenue venue.
(2) \emph{Within-author}: for authors with publications in both \nlpvenue and \mlvenue, we match each  author's \mlvenue     paper to that same author's own nearest \nlpvenue paper, and add     author fixed effects. The model is fit only from the papers of the 4,015  authors who publish at both venue families. Because a single same-author match is  looser than the corpus-wide one, we also add $k{=}50$ SPECTER2 cluster and year fixed effects to absorb residual topic differences.
The two regressions bound the role of author selection. One potential explanation is that researchers who publish at \mlvenue venues are systematically higher-ceiling to begin with (more coauthors, bigger followings, more cited in general), so their papers get more  citations regardless of venue. The within-author model tests this by estimating the premium with author identity held fixed; if the premium were due to higher- or lower-impact authors sorting by venue, the gap would collapse in this model.

\begin{table}[t]
  \centering
  \setlength{\tabcolsep}{4pt}
  \resizebox{0.49\textwidth}{!}{
  \begin{tabular}{@{}l r@{}l r@{}l@{}}
    \toprule
    & \multicolumn{2}{c}{Within-author} & \multicolumn{2}{c}{All papers} \\
    \midrule
    \textsc{ML-general} premium & +0.777 & $^{***}$\,\tiny{(0.045)} & +0.557 & $^{***}$\,\tiny{(0.032)} \\
    Intercept & \multicolumn{2}{c}{---} & +3.377 & $^{***}$\,\tiny{(0.101)} \\
    \midrule
    \multicolumn{5}{@{}l}{\emph{Publication-year effects (ref.\ 2015)}} \\
    \quad 2016 & -0.034 & \,\tiny{(0.125)} & +0.173 & \,\tiny{(0.129)} \\
    \quad 2017 & +0.178 & \,\tiny{(0.125)} & +0.440 & $^{**}$\,\tiny{(0.143)} \\
    \quad 2018 & -0.125 & \,\tiny{(0.128)} & +0.363 & $^{**}$\,\tiny{(0.122)} \\
    \quad 2019 & -0.318 & $^{*}$\,\tiny{(0.136)} & +0.236 & $^{*}$\,\tiny{(0.117)} \\
    \quad 2020 & -0.572 & $^{***}$\,\tiny{(0.122)} & +0.074 & \,\tiny{(0.111)} \\
    \quad 2021 & -0.763 & $^{***}$\,\tiny{(0.150)} & -0.163 & \,\tiny{(0.111)} \\
    \quad 2022 & -1.358 & $^{***}$\,\tiny{(0.124)} & -0.548 & $^{***}$\,\tiny{(0.106)} \\
    \quad 2023 & -1.500 & $^{***}$\,\tiny{(0.117)} & -0.921 & $^{***}$\,\tiny{(0.107)} \\
    \quad 2024 & -1.851 & $^{***}$\,\tiny{(0.128)} & -1.162 & $^{***}$\,\tiny{(0.101)} \\
    \midrule
    Author fixed effects & \multicolumn{2}{c}{Yes} & \multicolumn{2}{c}{No} \\
    Cluster fixed effects ($k{=}50$) & \multicolumn{2}{c}{Yes} & \multicolumn{2}{c}{No} \\
    Observations & \multicolumn{2}{c}{27,284} & \multicolumn{2}{c}{127,456} \\
    $R^2$ & \multicolumn{2}{c}{0.215} & \multicolumn{2}{c}{0.174} \\
    \bottomrule
  \end{tabular}
  }
  \caption{Fits of the two pooled citation-premium regressions. Publication-year effects are relative to 2015; the $k{=}50$ cluster dummies are included but summarized here. SEs are in parentheses; $^{*}p{<}0.05$, $^{**}p{<}0.01$, $^{***}p{<}0.001$.}
  \label{tab:citation_premium}
\end{table}

\paragraph{The premium is large, and author selection does not explain it.}
Table~\ref{tab:citation_premium} reports the pooled estimates. Matched to their
content-nearest *ACL papers, \mlvenue papers earn $+0.557$ log units
more citations ($p$ effectively zero; $\approx\!+75\%$) across all papers. When
the comparison is held \emph{within author}---each switcher's own
\mlvenue paper against their own nearest \nlpvenue paper---the
premium is $+0.777$ ($\approx\!+118\%$). Holding author identity fixed does not
shrink the premium; if anything it is larger, so the gap is not an artifact of
higher-reach authors sorting into \mlvenue. The citation premium is a
content-conditional venue effect, not a property of who publishes where.

\section{Conclusion}

Where should NLP research be published? Over the past decade authors working in NLP have gained an expanded set of venues to choose from as research on LLMs has blurred the lines between NLP, ML, and AI. Across our analyses, we find a major shift is underway: established NLP authors are increasingly moving their work from *ACL venues to more general ML venues like ICLR and NeurIPS. Decomposing this movement, we find the gains at \mlvenue venues and the losses at \aivenue venues are driven largely by \emph{convention}---the same authors, working on the same topics, increasingly choosing \mlvenue venues---while the decline in \nlpvenue share also reflects a shift in the cohort's topic mix toward content that is more common at ML venues. The shift is also generational: new NLP researchers increasingly debut their work at \mlvenue venues rather than *ACL. And it is plausibly reinforced by a citation premium at general-ML venues, where content-matched work attracts more citations. Read through a science-of-science lens, the NLP-to-ML shift is one axis of the field-level mobility that researchers display over their careers \citep{zeng-etal-2019-switch,jia-etal-2017-quantifying}: not authors abandoning their topics so much as a community reassigning where those topics are published, with the next generation entering at the venues the rewards now favor.
Our findings inform open questions on the future of NLP: how *ACL venues might respond to the shift; how generational replacement interacts with the venue prestige hierarchy; and whether the citation premium is durable or an artifact of the LLM era's outsized attention to ML-general venues. 

\section*{Limitations}
\label{sec:limitations}

\paragraph{Topic-judge precision is asymmetric across venues.} The
venue-blind LLM judge (Table~\ref{tab:judge_eval}) is most accurate on
*ACL papers (precision $0.94$, recall $0.98$) and slightly weaker
where NLP-topic papers are rare, with precision and recall of $0.91$
and $0.88$ at AI-broad venues and $0.79$ and $0.92$ at ML-general
venues. The cohort filters in
\S\ref{sec:migration_setup} and \S\ref{sec:phd_debut} rely on
baseline-window labels where high recall protects against
attriting genuine NLP authors, but post-period NLP-topic counts at
the rare-positive outcome categories are inflated by the residual
over-firing (lower precision) at ML-general venues. Approximately $9\%$ of the
cohort papers lack an abstract and fall back to a title-only judge
call.

\paragraph{Structural venue changes.} 

Multiple venues with no pre-LLM-era equivalent appear in the post window. Most notably, the *ACL ``Findings'' venue accounts for a substantial volume of papers, and newer ML-general venues such as COLM (2024) and TMLR (2022) also appear. These are sources of asymmetric supply expansion within our NLP and ML-general categories. The Findings expansion biases the post-2020 NLP supply upward (more *ACL slots, especially relevant for the trajectory in \S\ref{sec:migration_trajectory}); the COLM and TMLR launches and the NeurIPS D\&B track bias the post-2022 ML-general supply upward. The sliding-cutoff robustness (Appendix~\ref{sec:appendix_robustness}) bounds the
sensitivity to these structural changes.

\paragraph{PhD Cohort Platform Bias.} 

The PhD-student cohort is constructed primarily from OpenReview profiles, which over-represents authors who have submitted to OpenReview-managed venues (ICLR, NeurIPS, COLM, TMLR). Students whose first first-author submissions are exclusively to *ACL before it began switching to OpenReview in 2021, or to non-OpenReview-managed venues, are under-represented in our cohort. The DBLP and ORCID-education supplement we include mitigates but does not eliminate this bias; the magnitude of the NLP-debut shift may therefore be biased toward smaller values.

\paragraph{Causal vs.\ Descriptive Analysis.}
This paper documents empirical patterns of scholar behavior. The mechanisms behind those behaviors are complex and not easily or precisely quantified through causal analysis. While we use the agentic verb ``migrate,'' we are only  observing venue-share movement, not author intent. Further, while we have adopted matching procedures from causal inference, we do not make a causal claim that publishing a paper in a \mlvenue venue will increase citation counts. 

\section*{Ethics}
\label{sec:ethics}

This work analyzes aggregated bibliometric data about publication venues, authorship, and citation counts of academic papers. All data is derived from public sources and we do not identify individual researchers in our claims. Individual paper records were used solely for content-matching (SPECTER2 embeddings) and aggregate-statistics construction.

All public data was collected from sources designed for open access through official releases or their API: ACL Anthology, Semantic Scholar (under authenticated API terms), OpenAlex, OpenReview public profile/paper records, PMLR proceedings, DBLP bulk export, and ORCID educations where the researcher elected to make them public.

\ifblind
\else
\section*{Acknowledgments}
The author would like to especially thank all the researchers and engineering teams behind the ACL Anthology, Semantic Scholar, OpenAlex, OpenReview, and DBLP. Maintaining these systems is real work---especially with the growing number of papers and submissions---and the effort to make this data open enables studies like this one.
\fi

\bibliography{custom}

\appendix

\section{Supplemental analyses}
\label{sec:appendix}

This appendix gathers the material referenced from the main
text: further detail on the data (\S\ref{sec:appendix_data_details}),
robustness checks on the migration result (\S\ref{sec:appendix_robustness}),
and additional mechanism analyses (\S\ref{sec:appendix_mechanism}).

\subsection{Data Details}
\label{sec:appendix_data_details}

Because the cross-linked author graph underpins every cohort, we report how often each external identifier resolves. Table~\ref{tab:id_coverage_nlp} gives the share of authors carrying each handle: a Semantic Scholar identifier is present for nearly all authors, while OpenReview, DBLP, ORCID, and Google Scholar handles are progressively sparser.

\begin{table}[t]
  \centering
  \small
  \begin{tabular}{lrr}
    \toprule
    Identifier & Authors covered & Coverage \\
    \midrule
    Semantic Scholar & 130,780 & 97.0\% \\
    OpenReview & 35,757 & 26.5\% \\
    DBLP & 31,104 & 23.1\% \\
    ORCID & 23,871 & 17.7\% \\
    ACL Anthology & 13,787 & 10.2\% \\
    Google Scholar & 15,244 & 11.3\% \\
    Homepage URL & 14,370 & 10.7\% \\
    \bottomrule
  \end{tabular}
  \caption{Cross-link coverage for authors of $\geq$1 NLP-topic paper ($N=134,891$ of 320,775 total authors; venue-blind LLM topic label). This is the population the study analyzes; coverage rates exclude corpus co-authors who never published an NLP-topic paper.}
  \label{tab:id_coverage_nlp}
\end{table}

Table~\ref{tab:venue_taxonomy} lists the full venue catalog and its assignment to the \nlpvenue, \mlvenue, and \aivenue families, and Table~\ref{tab:per_venue_year_counts} reports per-venue paper counts across the three windows used throughout (2010--14, 2015--20, 2021--26). Coverage is dense from 2015 onward for every tracked venue.

\begin{table*}[t]
  \centering
  \small
  \begin{tabular}{lll}
    \toprule
    Category & Venue & Years \\
    \midrule
    AI-broad & AAAI & 1980--present \\
    AI-broad & IJCAI & 1969--present \\
    ML-general & COLM & 2024--present \\
    ML-general & ICLR & 2013--present \\
    ML-general & ICML & 1980--present \\
    ML-general & NeurIPS & 1987--present \\
    ML-general & TMLR & 2022--present \\
    NLP & AACL & 2020--present \\
    NLP & ACL Findings & 2021--present \\
    NLP & ACL (Main) & 1979--present \\
    NLP & Computational Linguistics (journal) & 1974--present \\
    NLP & COLING & 1965--present \\
    NLP & CoNLL & 1997--present \\
    NLP & EACL Findings & 2023--present \\
    NLP & EACL (Main) & 1983--present \\
    NLP & EMNLP Findings & 2020--present \\
    NLP & EMNLP (Main) & 1996--present \\
    NLP & IJCNLP & 2004--2023 \\
    NLP & LREC & 1998--present \\
    NLP & NAACL Findings & 2022--present \\
    NLP & NAACL (Main) & 2000--present \\
    NLP & Transactions of the ACL (TACL) & 2013--present \\
    NLP & WMT & 2006--present \\
    \bottomrule
  \end{tabular}
  \caption{Venue taxonomy used throughout the paper. The three tracked categories cover *ACL (NLP), the main ML-general venues (NeurIPS, ICLR, ICML, COLM, TMLR), and AAAI/IJCAI (AI-broad). The \emph{Years} column is each venue's active span (launch year to present, or to its final year); the study analyzes papers published from 2010 onward, with the primary analyses focused on 2015--2026.}
  \label{tab:venue_taxonomy}
\end{table*}

\begin{table*}[t]
  \centering
  \small
  \begin{tabular}{lrrrr}
    \toprule
    Venue & 2010-14 & 2015-20 & 2021-26 & Total \\
    \midrule
    NeurIPS & 1,863 & 5,985 & 18,535 & 26,383 \\
    AAAI & 1,904 & 6,454 & 16,877 & 25,235 \\
    ICML & 1,238 & 3,502 & 10,183 & 14,923 \\
    ICLR & 0 & 1,777 & 9,553 & 11,330 \\
    IJCAI & 494 & 4,710 & 4,756 & 9,960 \\
    EMNLP (Main) & 852 & 2,950 & 5,799 & 9,601 \\
    ACL (Main) & 1,323 & 2,827 & 5,124 & 9,274 \\
    LREC & 2,061 & 2,368 & 2,358 & 6,787 \\
    EMNLP Findings & 0 & 447 & 4,439 & 4,886 \\
    ACL Findings & 0 & 0 & 4,051 & 4,051 \\
    TMLR & 0 & 0 & 3,820 & 3,820 \\
    NAACL (Main) & 420 & 1,188 & 2,199 & 3,807 \\
    COLING & 914 & 1,378 & 1,389 & 3,681 \\
    EACL (Main) & 210 & 239 & 1,274 & 1,723 \\
    NAACL Findings & 0 & 0 & 980 & 980 \\
    IJCNLP & 391 & 179 & 329 & 899 \\
    TACL & 78 & 265 & 462 & 805 \\
    WMT & 0 & 141 & 591 & 732 \\
    COLM & 0 & 0 & 717 & 717 \\
    EACL Findings & 0 & 0 & 709 & 709 \\
    CoNLL & 0 & 411 & 200 & 611 \\
    Computational Linguistics & 205 & 174 & 178 & 557 \\
    AACL & 0 & 92 & 147 & 239 \\
    Total & 11,953 & 35,087 & 94,670 & 141,710 \\
    \bottomrule
  \end{tabular}
  \caption{Papers per venue, bucketed into three windows. Counts reflect the deduplicated union of the ACL Anthology, Semantic Scholar, OpenAlex, OpenReview, PMLR, and DBLP sources.}
  \label{tab:per_venue_year_counts}
\end{table*}

\begin{figure*}[t]
  \centering
  \includegraphics[width=0.95\textwidth]{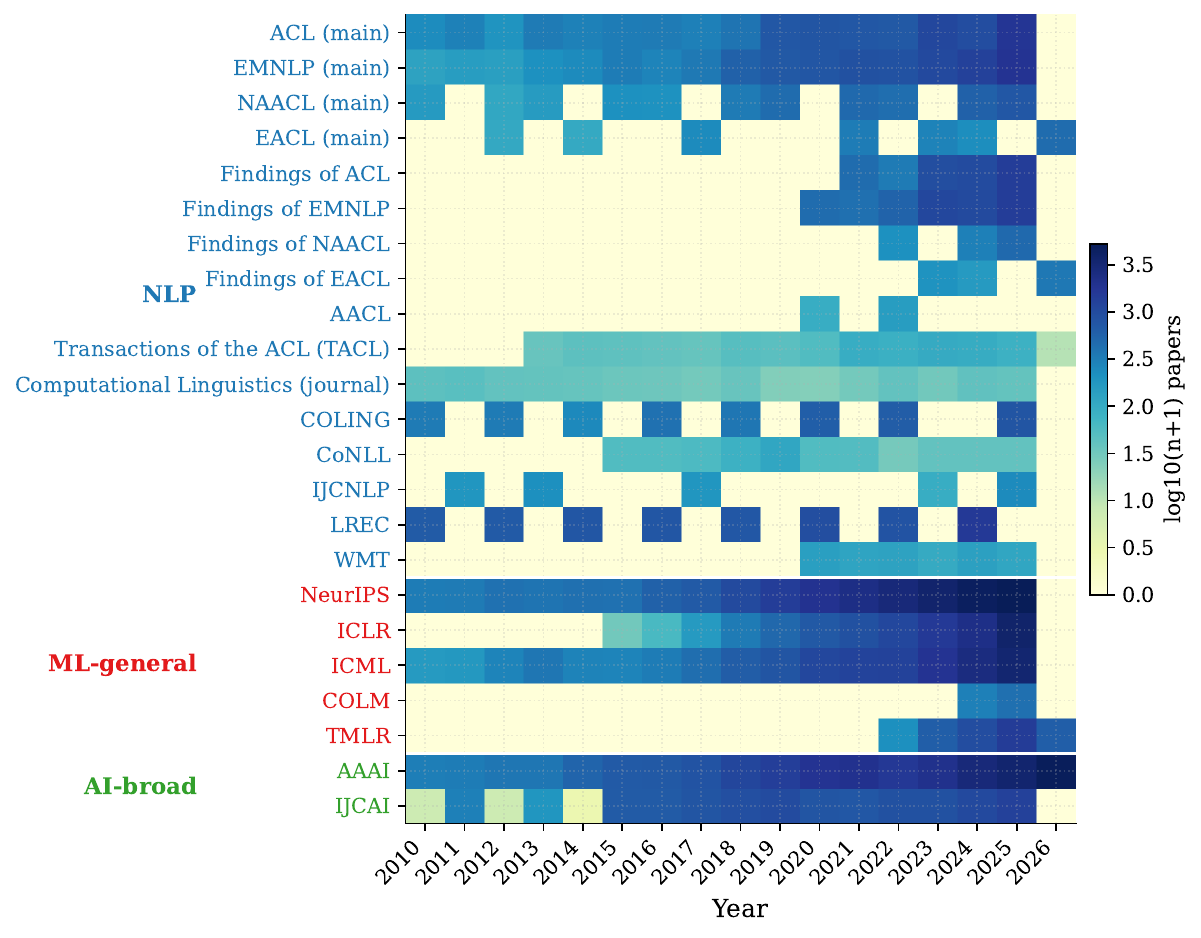}
  \caption{Per-venue/year paper coverage on a log color scale.
    Coverage is dense across the 2015--2026 window for all venues; the
    extension back to 2010 (used by \S\ref{sec:migration} for
    pre-LLM-era baseline) is solid for AAAI, IJCAI, NeurIPS, ICML, and
    the *ACL family. ICLR coverage begins in 2015 (its 2013--14
    editions were not backfilled, but they precede every analysis
    window); COLM launches in 2024.}
  \label{fig:data_coverage_heatmap}
\end{figure*}

The citation-premium analysis (\S\ref{sec:citation_premium}) relies on paper abstracts (for the SPECTER2 embeddings) and Semantic Scholar citation counts; both are available for the large majority of papers in all three families (Table~\ref{tab:abstract_citation_coverage}).

\begin{table}[t]
  \centering
  \small
  \begin{tabular}{lrrr}
    \toprule
    Category & Papers & Abstract \% & Citation \% \\
    \midrule
    NLP & 49,342 & 91.9 & 94.2 \\
    ML-general & 57,173 & 87.7 & 89.2 \\
    AI-broad & 35,195 & 94.3 & 88.8 \\
    Total & 141,710 & 90.8 & 90.8 \\
    \bottomrule
  \end{tabular}
  \caption{Abstract and citation coverage by venue category. Citation counts come from Semantic Scholar; abstracts come from Semantic Scholar, OpenAlex (inverted-index reconstruction), the OpenReview, PMLR, and ACL Anthology source records, and a tiered Crossref and publisher-landing-page scrape for IEEE/ACM/Springer DOIs.}
  \label{tab:abstract_citation_coverage}
\end{table}

The pool of $141{,}710$ distinct papers is assembled from complementary sources, deduplicated to a canonical paper identifier (DOI $>$ Semantic Scholar $>$ OpenAlex $>$ ACL Anthology $>$ DBLP $>$ arXiv $>$ OpenReview): the ACL Anthology ($49{,}342$ papers; *ACL ground truth), DBLP venue extracts ($62{,}172$; the primary source for AAAI/IJCAI and much of NeurIPS/ICLR where Anthology coverage is absent), PMLR ($14{,}199$; ICML), OpenReview ($10{,}946$; NeurIPS/ICLR/COLM/TMLR), and Semantic Scholar's venue-paper endpoint as a backfill ($5{,}051$). Almost all papers carry a Semantic Scholar identifier, though DOIs are sparse for the ML-general venues (which are ingested from OpenReview/PMLR without DOIs).

Table~\ref{tab:exp1_funnel} traces the established-author cohort from the full author pool down to the 4,181 research-active authors we study.

\begin{table}[t]
  \centering 
  \resizebox{0.48\textwidth}{!}{
  \begin{tabular}{lr}
    \toprule
    Filter & Authors \\
    \midrule
    All corpus authors (population) & 320{,}775 \\
    \quad + $\ge 1$ tracked-family paper in 2015--2020 baseline & 50{,}552 \\
    \quad + $\ge 3$ NLP-topic papers in baseline & \\
    \quad\quad AND $\ge 50\%$ NLP-topic share in baseline & 5{,}403 \\
    \quad + $\ge 1$ NLP-topic paper in 2021--26 (active) & \textbf{4{,}181} \\
    \bottomrule
  \end{tabular}
  }
  \caption{Established-author cohort selection funnel, using the venue-blind LLM topic label (\S\ref{sec:appendix_prompt}).}
  \label{tab:exp1_funnel}
\end{table}

\subsubsection{New Entrant Cohort Identification}
\label{sec:appendix_phd_cohort}

\begin{table*}[t]
  \centering
  \small
  \begin{tabular}{lrp{0.55\linewidth}}
    \toprule
    Cohort & $N$ & Definition \\
    \midrule
    Established authors: topic-active & 4,181 & $\ge 3$ NLP-topic papers in 2015--20, $\ge 50\%$ NLP share, $\ge 1$ NLP-topic paper in 2021--26 \\
    Established authors: no activity floor (robustness) & 5,403 & $\ge 3$ NLP-topic papers in 2015--20, $\ge 50\%$ NLP share; drops the $\ge 1$ post-period NLP-topic paper requirement \\
    New entrants: publication-record & 3,568 & $\ge 3$ first-author NLP-topic papers, first one in 2019--2024 (entry = year of first NLP-topic first-author paper) \\
    New entrants: all PhD students & 25,521 & PhD start 2019--2024 from OpenReview $>$ ORCID $>$ DBLP-thesis backfill (full) \\
    New entrants: declared-PhD & 1,228 & PhD start 2019--2024 \emph{and} $\ge 3$ first-author NLP-topic papers \\
    \bottomrule
  \end{tabular}
  \caption{Cohort definitions and sizes. Topic-active is the primary established-author cohort; the no-activity-floor variant is reported in the appendix as a robustness check. The two new-entrant cohorts (publication-record and declared-PhD) each restrict to new entrants with $\ge 3$ first-author NLP-topic papers; the all-PhD-students row is the pre-restriction declared-PhD population.}
  \label{tab:cohort_sizes}
\end{table*}

\paragraph{Publication-record Cohort.}

For every author in our corpus we identify their first-author (position $=0$) NLP-topic papers at the three tracked venue families. The publication-record cohort consists of every author with at least three such papers whose first one falls in $[2019, 2024]$; the entry (cohort) year is the year of that first NLP-topic first-author paper ($N = 3{,}568$ after author disambiguation). Widening the entry window to $[2017, 2024]$ yields $N = 4{,}259$; Table~\ref{tab:exp2_unified} reports the $[2019, 2024]$ window.

This definition removes the platform-coverage selection that affects
the PhD-student cohort: every author who accumulates three first-author NLP-topic papers in the corpus is included, regardless of whether they have an OpenReview, ORCID, or DBLP-thesis profile.

\textbf{Declared-PhD Cohort.} 
A researcher enters the declared-PhD cohort iff all four criteria below hold. The criteria are applied in order; (C1)--(C3) build the \emph{all-PhD-students} cohort and (C4) restricts it to the cohort used in experiments.

\begin{enumerate}
  \item[\textbf{C1}] \textbf{A PhD-equivalent education record.}
    Detected from a three-source resolver (OpenReview, ORCID, and DBLP-thesis records) applied in strict priority (a higher source wins; a lower one only adds researchers not already matched):
    \emph{(a)~OpenReview} through any \texttt{content.history} or     \texttt{content.education} entry whose \texttt{position} string     contains, case-insensitively, one of \{\texttt{phd}, \texttt{ph.d}, \texttt{doctoral}, \texttt{doctorate}\}; 
    \emph{(b)~ORCID} through an \texttt{educations}      entry whose role title matches     \texttt{\textbackslash b(ph\textbackslash.?d|doctoral|doctorate)\textbackslash b} or contains ``phd student''/``phd candidate'';
    \emph{(c)~DBLP-thesis} when the author has a registry     \texttt{\textless phdthesis\textgreater} record, which is treated as a supplemental backfill only.
  
  \item[\textbf{C2}] \textbf{Resolvable to a unified author.} The profile links to an \texttt{author\_uid} in our author table (OpenReview via \texttt{openreview\_id}, ORCID via \texttt{orcid}, DBLP via the thesis record's \texttt{author\_uid}); profiles that do not link are dropped.
  
  \item[\textbf{C3}] \textbf{PhD start year in 2019--2024.} An integer start year parsed from the education entry (OpenReview/ORCID: the self-reported education start; DBLP: inferred as $\text{dissertation\_year}-5$) falls in     $[2019,2024]$. When a researcher has several PhD entries (which is very rare) the qualifying one is used.
    
  \item[\textbf{C4}] \textbf{At least three first-author NLP-topic papers.} The author has $\ge 3$ papers as a first author on NLP topics at a tracked venue in 2019--2026; the first three by year are the analysis set.
\end{enumerate}

Note that the DBLP thesis records contain the \emph{completion} year rather than a start year; we infer a likely starting date for the PhD using the national average degree time in China and North America of 5 years.\footnote{Students from institutions in these regions account for the majority of the authors, though we recognize that European students often have a much shorter time to degree of $\sim$3 years.}

Criteria (C1)--(C3) yield the full PhD-student cohort (the all-PhD-students cohort); adding (C4) yields the declared-PhD cohort, which contains $N$=1,228 unique students (by resolving source: OpenReview 1,010, ORCID 150, DBLP-thesis backfill 68).
Table~\ref{tab:exp2_funnel} shows the step-by-step funnel from
population to the final cohort.

\begin{table}[t]
  \centering
  \small
  \begin{tabular}{lr}
    \toprule
    Filter & Students \\
    \midrule
    OpenReview profiles (population) & 84{,}708 \\
    \quad + parseable PhD start year & 51{,}288 \\
    \quad + PhD start 2019--2024 (OpenReview) & 27{,}067 \\
    \quad + $\ge 1$ tracked-cat first-author paper 2019--26, & \\
    \quad\quad unioned w/ ORCID + DBLP-thesis backfill & 25{,}521 \\
    \quad + $\ge 3$ first-author NLP-topic papers & \textbf{1{,}228} \\
    \bottomrule
  \end{tabular}
  \caption{Declared-PhD cohort selection funnel. The first four rows describe the all-PhD-students population (25{,}521 students with at least one first-author paper); the last row produces the declared-PhD cohort (1{,}228 students with $\ge 3$ first-author NLP-topic papers).}
  \label{tab:exp2_funnel}
\end{table}

\subsection{Robustness}
\label{sec:appendix_robustness}

Table~\ref{tab:exp1_stacked_headline} gives the full coefficients behind the forest plot in \S\ref{sec:migration_delta_share}: the mean within-author change in each venue family's share, under any-authorship and under first-or-last-authorship, with cluster-robust standard errors and a joint Wald test against the null of an unchanged venue mix.

\begin{table*}[t]
  \centering
  \small
  \begin{tabular}{l rrr rrr}
    \toprule
    Venue category & $\Delta$share (pp) & SE & $p_{\text{Holm}}$ & $\Delta$share (pp)  & SE  & $p_{\text{Holm}}$  \\
    \midrule
    \venue{*ACL} (NLP) & -8.96 & 0.57 & $<$0.001 & -14.18 & 0.78 & $<$0.001 \\
    \venue{ML-general} & +8.60 & 0.36 & $<$0.001 & +5.58 & 0.39 & $<$0.001 \\
    \venue{AI-broad} & -4.57 & 0.41 & $<$0.001 & -4.84 & 0.45 & $<$0.001 \\
    \bottomrule
  \end{tabular}
  \\\footnotesize{Estimator: stacked OLS $\Delta\text{share}_{ic} \sim 0 + C(\text{category})$, one row per (author, venue category), cluster-robust SE by author ($N=4,181$ authors). Each coefficient is the mean \emph{within-author} change in venue-mix share from 2015--20 to 2021--26, in percentage points. Left block: any-authorship; right block: first-or-last-author. Joint Wald test of $H_0$: venue mix unchanged --- any-authorship $p=8.9\times10^{-128}$, first/last $p=2.8\times10^{-99}$. Per-category $p_{\text{Holm}}$ is Holm--Bonferroni FWER-adjusted within each authorship column. Per-stratum coefficients are in Appendix Table~\ref{tab:exp1_stratum_combined}.}
  \caption{Per-author venue-mix shift (2015--20 $\to$ 2021--26). Any-authorship (left) and first-or-last-author (right).}
  \label{tab:exp1_stacked_headline}
\end{table*}

\paragraph{Is 2020 a reasonable pre/post cutoff?}
We split the study window at 2020 because it brackets the arrival of the first large language models (GPT-3), but the venue-mix shift should not depend on that exact year choice. We therefore recompute the per-author $\Delta$share while sliding the split year $k$ from 2017 to 2022, each time pairing an equal three-year baseline $[k{-}2,k]$ with a post window $[k{+}1,k{+}3]$ and rebuilding the cohort (Table~\ref{tab:sliding_cutoff}); this range brackets the inflection and uses a three-year window (rather than the six-year main windows) so that it is fully covered by the 2010--2025 data. The \nlpvenue decline and the \mlvenue rise appear at \emph{every} cutoff, so neither is an artifact of splitting at 2020. The \mlvenue gain also sharpens as the split moves later, growing monotonically from $+1.7$pp at $k{=}2017$ to $+9.0$pp at $k{=}2022$, while \aivenue stays negative throughout. This mirrors the trajectory inflection in \S\ref{sec:migration_trajectory} and places the acceleration in the post-2020 window, so 2020 is a reasonable boundary, if slightly conservative.

\begin{table}[t]
  \centering
  \small
  \begin{tabular}{c r r@{}l r@{}l r@{}l}
    \toprule
    Cutoff $k$ & $N$ & \multicolumn{2}{c}{\nlpvenue} & \multicolumn{2}{c}{\mlvenue} & \multicolumn{2}{c}{\aivenue} \\
    \midrule
    2017 & 1,812 & -17.2 & $^{*}$ & +1.7 & $^{*}$ & -2.8 & $^{*}$ \\
    2018 & 2,503 & -21.8 & $^{*}$ & +1.6 & $^{*}$ & -2.6 & $^{*}$ \\
    2019 & 2,648 & -14.7 & $^{*}$ & +2.4 & $^{*}$ & -5.4 & $^{*}$ \\
    2020 & 3,897 & -17.7 & $^{*}$ & +3.6 & $^{*}$ & -6.1 & $^{*}$ \\
    2021 & 4,594 & -19.1 & $^{*}$ & +6.3 & $^{*}$ & -7.3 & $^{*}$ \\
    2022 & 6,168 & -28.3 & $^{*}$ & +9.0 & $^{*}$ & -4.1 & $^{*}$ \\
    \bottomrule
  \end{tabular}
  \caption{Sliding pre/post-cutoff sensitivity. Each row rebuilds the cohort with an equal three-year baseline $[k{-}2,k]$ and post window $[k{+}1,k{+}3]$ and reports the mean within-author change in venue-mix share (percentage points), renormalized within the three tracked families. We sweep the split year $k$ across 2017--2022, the range that brackets the LLM inflection and that the 2010--2025 data fully support at this window width. The \nlpvenue decline and \mlvenue rise hold at every split year, and the \mlvenue gain grows monotonically with $k$ while \aivenue stays negative throughout, so neither trend is an artifact of splitting at any particular year. $^{*}$ marks a 95\% CI excluding zero.}
  \label{tab:sliding_cutoff}
\end{table}

\paragraph{Author heterogeneity.}
Table~\ref{tab:exp1_stratum_combined} re-fits the venue-mix shift within author strata---career age and $h$-index quartile, as two separate stacked regressions with Holm--Bonferroni--corrected $p$-values. The \nlpvenue decline and \mlvenue gain persist in every stratum, so the migration is not confined to any one seniority or impact group. Across career age no stratum deviation survives correction (the shift is essentially uniform); the one sizeable, correction-surviving exception is by citation impact (Panel B), where the highest-$h$-index authors retain \nlpvenue much more than the lowest. Paper-count quartile mirrors the $h$-index pattern more weakly and the hybrid career-$\times$-role split is too sparse to interpret, so both are omitted.

\begin{table*}[t]
  \centering
  \small
  \begin{tabular}{l rrr rrr}
    \toprule
    & \multicolumn{3}{c}{Any-authorship} & \multicolumn{3}{c}{First-or-last-author} \\
    \cmidrule(lr){2-4}\cmidrule(lr){5-7}
    Term & $\Delta$pp & SE & $p_{\text{Holm}}$ & $\Delta$pp & SE & $p_{\text{Holm}}$ \\
    \midrule
    \multicolumn{7}{l}{\textit{Panel A: career age (ref.\ junior $\le$3y)}} \\
    \venue{*ACL} (NLP) & -11.53 & 2.11 & $<$0.001 & -17.90 & 3.04 & $<$0.001 \\
    \venue{ML-general} & +7.37 & 1.19 & $<$0.001 & +3.33 & 1.22 & 0.066 \\
    \venue{AI-broad} & -1.32 & 1.54 & 1.000 & -5.10 & 1.59 & 0.015 \\
    \venue{*ACL} (NLP) $\times$ mid (4--8y) & +2.81 & 2.45 & 1.000 & +7.26 & 3.51 & 0.276 \\
    \venue{ML-general} $\times$ mid (4--8y) & +1.31 & 1.45 & 1.000 & +3.21 & 1.52 & 0.276 \\
    \venue{AI-broad} $\times$ mid (4--8y) & -2.99 & 1.81 & 0.687 & +1.31 & 1.88 & 1.000 \\
    \venue{*ACL} (NLP) $\times$ senior (9--15y) & +2.91 & 2.35 & 1.000 & +1.97 & 3.36 & 1.000 \\
    \venue{ML-general} $\times$ senior (9--15y) & +3.17 & 1.39 & 0.201 & +3.32 & 1.44 & 0.194 \\
    \venue{AI-broad} $\times$ senior (9--15y) & -4.58 & 1.72 & 0.077 & -1.23 & 1.82 & 1.000 \\
    \venue{*ACL} (NLP) $\times$ veteran ($>$15y) & +2.67 & 2.28 & 1.000 & +3.90 & 3.25 & 1.000 \\
    \venue{ML-general} $\times$ veteran ($>$15y) & -0.04 & 1.30 & 1.000 & +1.38 & 1.34 & 1.000 \\
    \venue{AI-broad} $\times$ veteran ($>$15y) & -2.97 & 1.65 & 0.574 & +0.87 & 1.72 & 1.000 \\
    \midrule
    \multicolumn{7}{l}{\textit{Panel B: $h$-index quartile (ref.\ Q1)}} \\
    \venue{*ACL} (NLP) & -13.01 & 1.13 & $<$0.001 & -18.77 & 1.57 & $<$0.001 \\
    \venue{ML-general} & +8.92 & 0.69 & $<$0.001 & +5.32 & 0.69 & $<$0.001 \\
    \venue{AI-broad} & -1.14 & 0.81 & 0.646 & -2.59 & 0.85 & 0.027 \\
    \venue{*ACL} (NLP) $\times$ Q2 & +3.46 & 1.67 & 0.247 & +5.00 & 2.32 & 0.219 \\
    \venue{ML-general} $\times$ Q2 & -0.20 & 1.05 & 1.000 & +0.28 & 1.09 & 1.000 \\
    \venue{AI-broad} $\times$ Q2 & -3.50 & 1.22 & 0.036 & -3.14 & 1.28 & 0.127 \\
    \venue{*ACL} (NLP) $\times$ Q3 & +4.88 & 1.58 & 0.021 & +3.42 & 2.21 & 0.733 \\
    \venue{ML-general} $\times$ Q3 & -0.32 & 1.00 & 1.000 & -0.16 & 1.04 & 1.000 \\
    \venue{AI-broad} $\times$ Q3 & -4.36 & 1.14 & 0.001 & -2.97 & 1.27 & 0.152 \\
    \venue{*ACL} (NLP) $\times$ Q4 & +8.36 & 1.55 & $<$0.001 & +10.69 & 2.06 & $<$0.001 \\
    \venue{ML-general} $\times$ Q4 & -0.80 & 1.00 & 1.000 & +1.01 & 1.05 & 1.000 \\
    \venue{AI-broad} $\times$ Q4 & -6.28 & 1.10 & $<$0.001 & -3.16 & 1.19 & 0.080 \\
    \venue{*ACL} (NLP) $\times$ unknown $h$ & +23.97 & 10.39 & 0.169 & +30.90 & 10.92 & 0.051 \\
    \venue{ML-general} $\times$ unknown $h$ & -6.06 & 2.88 & 0.247 & +0.57 & 4.20 & 1.000 \\
    \venue{AI-broad} $\times$ unknown $h$ & -12.68 & 8.36 & 0.646 & -15.43 & 10.67 & 0.741 \\
    \bottomrule
  \end{tabular}
  \caption{Author heterogeneity of the venue-mix shift (2015--20 $\to$ 2021--26), in percentage points. Panels A (top) and B (bottom) are \emph{two separate} stacked OLS regressions ($\Delta\text{share} \sim 0 + C(\text{category}) + C(\text{category}){:}C(\text{stratum})$), each with its own reference stratum; the rows without $\times$ are the reference-stratum category means and the remaining rows are stratum-level deviations. All $p$-values are Holm--Bonferroni corrected within each panel and authorship column. The \nlpvenue decline and \mlvenue gain hold in every stratum; the only sizeable, correction-surviving heterogeneity is by $h$-index (Panel B): the highest-impact authors retain \nlpvenue venues the most.}
  \label{tab:exp1_stratum_combined}
\end{table*}

\subsection{Mechanism supplements}
\label{sec:appendix_mechanism}

\begin{figure*}[t]
  \centering
  \includegraphics[width=0.95\textwidth]{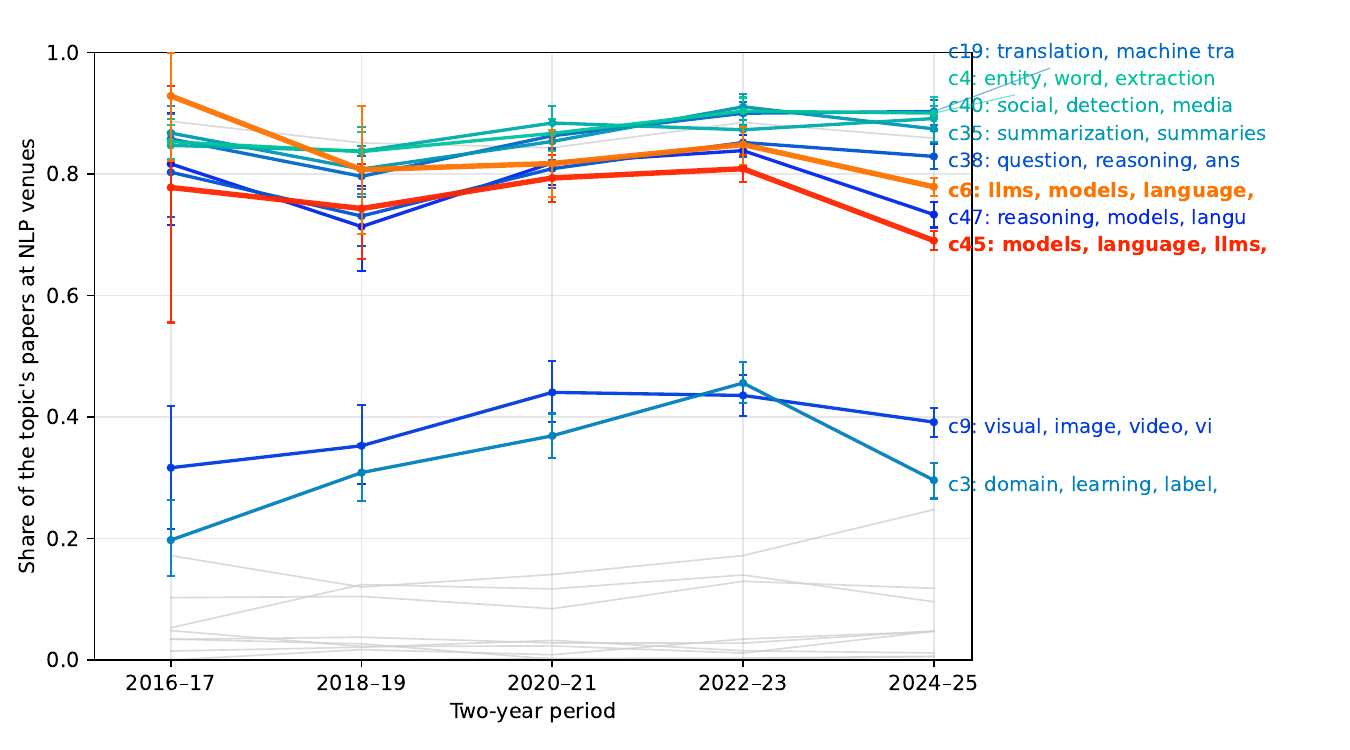}
  \caption{Share of each topic cluster's papers that appear at \nlpvenue
    venues, pooled into two-year periods (2016--2025; the partial 2026 is
    dropped, and binning removes the even/odd sawtooth created by biennial and
    cyclic *ACL venues such as LREC, EACL, NAACL, and COLING). We highlight
    the ten largest clusters with a substantial \nlpvenue presence. The two
    large language-model clusters (c45: language models/LLMs; c6: LLM
    evaluation) are the only ones whose \nlpvenue share \emph{falls} (reddish colors,
    bold)---these are the topics migrating to \mlvenue venues---while classic
    NLP tasks such as machine translation, question answering, summarization,
    and information extraction hold or consolidate at \nlpvenue (bluish colors). Gray
    lines are the remaining clusters. Error bars are bootstrapped 95\% CIs.}
  \label{fig:exp15_topic_time_series}
\end{figure*}

Table~\ref{tab:exp15_decomposition} gives the full Oaxaca--Blinder decomposition summarized in \S\ref{sec:mechanism_decomposition}, splitting each family's cohort $\Delta$share into a composition term (the cohort changed which topics it works on) and a convention term (the field changed where a given topic is published), under both the OpenAlex-subfield and the $k{=}50$ cluster taxonomies. As in the main text, the \mlvenue gain and \aivenue loss are convention-dominated, while the \nlpvenue decline is carried by composition. Figure~\ref{fig:exp15_topic_time_series} shows this at the topic level as the within-topic \nlpvenue share over time for the most common subfields, where a subfield whose share falls contributes negative convention.

\begin{table*}[t]
  \centering
  \small
  \begin{tabular}{lrrrrr}
    \toprule
    Taxonomy & Category & $\Delta$share & Composition & Convention & Interaction \\
    \midrule
    OpenAlex subfield & NLP & -0.014 & -0.034 & +0.020 & +0.000 \\
    OpenAlex subfield & ML-general & +0.080 & +0.033 & +0.047 & +0.000 \\
    OpenAlex subfield & AI-broad & -0.065 & +0.001 & -0.066 & +0.000 \\
    $k{=}50$ SPECTER2 cluster & NLP & -0.083 & -0.083 & -0.000 & -0.000 \\
    $k{=}50$ SPECTER2 cluster & ML-general & +0.119 & +0.061 & +0.058 & +0.000 \\
    $k{=}50$ SPECTER2 cluster & AI-broad & -0.036 & +0.022 & -0.058 & +0.000 \\
    \bottomrule
  \end{tabular}
  \caption{Oaxaca--Blinder decomposition of cohort $\Delta$share (post 2021--26 vs.\ baseline 2015--20) into composition (topic-mix shift, H1) and convention (within-topic venue shift, H2) under two taxonomies. Average of pre/post weight references.}
  \label{tab:exp15_decomposition}
\end{table*}

\paragraph{Citation premium: observable controls and author fixed effects.}
We refit the citation premium two further ways: (i) an observable-controls
regression with a venue$\times$OpenAlex-field interaction, paper age, and a
first/last-author flag, and (ii) an author fixed-effects spec that
identifies off authors who publish in more than one venue category.
\mlvenue venues lack OpenAlex topics (they have no DOI match),
so we backfill their field labels from Semantic Scholar fields-of-study
(overwhelmingly Computer Science), which makes the
venue$\times$field interaction identifiable.

\begin{figure}[t]
  \centering
  \includegraphics[width=\columnwidth]{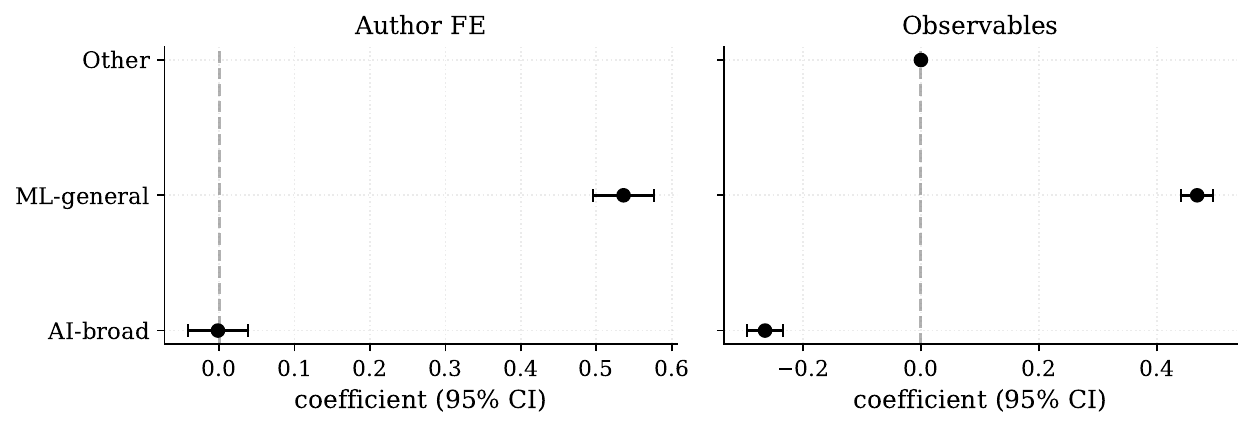}
  \caption{Citation premium (vs.\ *ACL/NLP) by venue category under the
    observable-controls and author-fixed-effects specifications. The
    \mlvenue premium is large and positive under both; the
    within-author estimate is close to the between-author one, indicating
    little author-selection bias.}
  \label{fig:citation_premium_forest}
\end{figure}

For Computer Science papers (the bulk of the cohort), the
observable-controls \mlvenue premium is $+0.47$ log units
($\approx$60\%; $p<10^{-200}$), and the within-author premium is $+0.54$
($\approx$71\%; $95\%$ CI $[0.50,0.58]$, $p<10^{-100}$). The within-author
estimate closely tracks the $+0.557$ all-papers matched estimate
(Table~\ref{tab:citation_premium}), so holding
author identity fixed barely changes the premium. As a result, author selection by venue
explains little of the citation gap.

\clearpage
\subsection{NLP-topic judge: full prompt}
\label{sec:appendix_prompt}

The NLP-topic label (\S\ref{sec:data_topics}) is produced by
Gemma-4-26B-A4B-it using a judge prompt grounded in calls for papers:
a call-for-papers area union (2015--2024), hardened
out-of-scope negatives and decision rules, and 16 few-shot examples,
designed to be applied venue-blind without an external
*ACL-membership override.
The prompt is \emph{venue-blind}: the model sees only the paper title
and abstract, never the venue. It is reproduced below verbatim---
\texttt{\{TITLE\}} and \texttt{\{ABSTRACT\}} are substituted with the
paper's text at inference time, the model is queried at temperature~$0$,
and a regular expression parses the trailing
\texttt{LABEL:\ YES}/\texttt{NO} from its single-line reply. Em- and
en-dashes are shown as \texttt{-{}-}/\texttt{-} for display; the prompt
is otherwise identical to the deployed template.

\begin{lstlisting}[style=prompt]
TASK: Decide whether a paper's PRIMARY research contribution falls within
Natural Language Processing / Computational Linguistics, defined as the
topical scope of the main ACL, EMNLP, and NAACL conferences (the ACL
Anthology core), unioned across the 2015-2024 Call-for-Papers area sets.
Use the topic, not where it was published: a paper is in scope if it would
naturally fit one of the ACL/EMNLP/NAACL area tracks below, whatever venue
it actually appeared in.

IN-SCOPE AREAS (ACL/EMNLP/NAACL Call-for-Papers areas, 2015-2024 -- the set has
grown over time; all of the following count, old and new):
- Syntax: tagging, chunking, parsing; phonology, morphology, word segmentation
- Semantics (lexical, sentence-level, textual inference); discourse, coreference, pragmatics
- Machine translation; multilinguality and language diversity
- Information extraction; text mining / IR as language understanding; question answering
- Summarization; natural language generation
- Dialogue and interactive systems
- Sentiment analysis, stylistic analysis, argument mining
- Language resources, datasets, evaluation and benchmarking for language tasks
- Machine learning / statistical methods FOR NLP; representation learning and language models studied as language systems
- Interpretability and analysis of NLP models; ethics, bias, fairness in NLP
- Efficient / low-resource methods for NLP
- Multimodality and language grounding to vision/robotics; language and vision
- Speech / spoken language understanding where language (not just acoustics) is central
- Computational social science / cultural analytics via text; NLP for web and social media; linguistic, psycholinguistic and cognitive modeling of language
- NLP applications (clinical, legal, scientific, educational, ...) where the contribution is the language method
- Large-language-model work IS in scope when the contribution concerns language (capabilities, behavior, training/evaluation, analysis, generation, multilingual or reasoning-in-language aspects)
- Instruction tuning, RLHF, alignment, agentic LLMs, tool use -- in scope when the contribution is a language capability
- Inference/training efficiency for LLMs and language models -- in scope (the model studied IS a language model)


OUT OF SCOPE (answer NO even if text, an LLM, or a linguistic term appears):
- A general machine-learning / optimization / statistics / theory contribution whose novelty is the ML method itself, merely demonstrated on a text dataset, or that lists text as one of several application domains (the typical NeurIPS/ICLR/ICML methods paper).
- Computer vision / image / video where text or captions are incidental, and tasks where natural language is only an instruction or control interface for a non-language goal (image or chart editing, GUI control, robot actuation).
- Information-retrieval, recommender, database or data-mining systems work centered on indexing, ranking, scalability, click-through or efficiency rather than language understanding.
- SPEECH that is acoustic / signal processing: text-to-speech and speech synthesis, voice conversion, vocoders, speaker verification / identification / diarization, anti-spoofing, audio enhancement -- the contribution there is audio, not language. (ASR, spoken-language understanding, spoken QA / dialogue, speech translation, phonology, word segmentation, and spoken language identification ARE in scope: language is central there.)
- A paper that USES a large language model or LLM agent merely as a tool or component, while its actual task and contribution is non-linguistic (recommendation, point-of-interest or trajectory prediction, forecasting, tabular prediction, numeric regression, planning, optimization, scientific discovery).
- Robotics, networks, security, HCI, bioinformatics, etc., where language is not the central object.

DECISION RULES:
1. Ask "is the central novelty a language / linguistic capability or understanding (YES), or a general method / another field that merely uses text or an LLM (NO)?"
2. A method whose contribution is APPLIED TO language is YES (instruction tuning of an LLM, RL for reasoning expressed in language, efficiency for a language model, a dataset or benchmark for an NLP task). But a generic method that merely runs on a text dataset, and a paper that uses an LLM to perform a non-language task, are NO -- judge what the contribution IS, not which tools it uses.
3. Vision or speech with language: YES only if a language / linguistic capability is the central contribution; NO if language is just a conditioning signal or instruction interface, or if the contribution is acoustic, visual, or systems.
4. Multilingual, low-resource, and applied NLP (clinical, legal, educational, social-media) -- YES when the language method, linguistic analysis, or language resource is the contribution.
5. Do NOT default to YES when uncertain: if you cannot identify a concrete language / linguistic contribution, answer NO. Judge ONLY from the title and abstract; if genuinely mixed, decide by the PRIMARY contribution.


EXAMPLES:

Example 1:
Title: Neural Machine Translation by Jointly Learning to Align and Translate
Abstract: We propose a neural encoder-decoder for translation with an attention mechanism that aligns source and target words.
REASON: Machine translation, a core NLP area | LABEL: YES

Example 2:
Title: Sharper Generalization Bounds for Stochastic Gradient Descent
Abstract: We derive new high-probability generalization bounds for SGD on convex losses; experiments include a text classification benchmark.
REASON: General ML theory -- text dataset is incidental | LABEL: NO

Example 3:
Title: Visual Question Answering with Question-Aware Image Features
Abstract: We study VQA, learning a multimodal representation that grounds natural-language questions in images and generates a free-form answer.
REASON: Language grounding to vision -- multimodal NLP | LABEL: YES

Example 4:
Title: Improving Naturalness in Unit-Selection Text-to-Speech with Neural Acoustic Embeddings
Abstract: We learn acoustic unit embeddings that improve waveform concatenation, raising the naturalness mean-opinion-score of a text-to-speech system.
REASON: Text-to-speech -- contribution is acoustic naturalness, not language | LABEL: NO

Example 5:
Title: End-to-End Speech Recognition for Four Low-Resource Bantu Languages
Abstract: We build an end-to-end ASR system for under-resourced Bantu languages and study cross-lingual transfer of lexical and acoustic units.
REASON: Speech recognition / spoken-language understanding -- language is central | LABEL: YES

Example 6:
Title: Self-Supervised Pretraining for Object Detection
Abstract: We pretrain a backbone on unlabeled images using contrastive learning and fine-tune for object detection; image captions provide weak supervision.
REASON: Vision contribution -- captions are incidental | LABEL: NO

Example 7:
Title: How Do In-Context Examples Affect LLM Reasoning?
Abstract: We probe instruction-tuned LLMs across reasoning benchmarks and analyze sensitivity to example order, demonstrating systematic effects.
REASON: LLM analysis -- language-model behavior | LABEL: YES

Example 8:
Title: An LLM-Agent Framework for Next Point-of-Interest Recommendation
Abstract: We orchestrate an LLM agent over user check-in trajectories to predict the next point of interest, improving recommendation accuracy on mobility data.
REASON: LLM is only a tool -- the task is location recommendation, not language | LABEL: NO

Example 9:
Title: Reinforcement Learning from Human Feedback for Open-Ended Generation
Abstract: We train a reward model from pairwise human preferences and fine-tune an LLM via PPO, improving helpfulness and safety of generation.
REASON: RLHF for language generation | LABEL: YES

Example 10:
Title: Scalable Index Compression for Web-Scale Retrieval
Abstract: We propose a compressed inverted index with delta encoding that reduces storage 3x and improves query throughput on web search.
REASON: IR systems / indexing -- not language understanding | LABEL: NO

Example 11:
Title: Detecting Community Mental-Health Signals from Longitudinal Social-Media Posts
Abstract: We model the language of users' social-media timelines to track community-level mental-health signals over time.
REASON: Computational social science via social-media text | LABEL: YES

Example 12:
Title: Robust Speaker Verification with Anti-Spoofing Countermeasures
Abstract: We propose acoustic countermeasures that make speaker verification robust to replay and synthetic-speech spoofing attacks.
REASON: Acoustic / biometric speech -- no language content | LABEL: NO

Example 13:
Title: A Multilingual Benchmark for Cross-Lingual Question Answering
Abstract: We release a 12-language QA benchmark with native-speaker annotations and evaluate state-of-the-art multilingual encoders.
REASON: Language resource / benchmark for NLP | LABEL: YES

Example 14:
Title: Refining Weak-Supervision Labeling Functions with Limited Labeled Data
Abstract: We propose a method that iteratively refines labeling functions for programmatic weak supervision, evaluated on text, tabular, and image benchmarks.
REASON: Generic ML method -- text is one of several application domains | LABEL: NO

Example 15:
Title: Extracting Medication Events from Clinical Notes with a Span-Based Tagger
Abstract: We present a span-based information-extraction model that identifies medications, dosages, and temporal relations in free-text clinical notes.
REASON: Clinical NLP -- information extraction is the language contribution | LABEL: YES

Example 16:
Title: Click-Through-Rate Prediction with Pretrained Text Features for E-Commerce
Abstract: We feed pretrained embeddings of product descriptions into a click-through-rate model, improving recommendation ranking.
REASON: Recommendation task -- text is an incidental input feature | LABEL: NO


Answer with EXACTLY one line, no extra text, in this format:
REASON: <=12 words | LABEL: YES or NO

Title: {TITLE}
Abstract: {ABSTRACT}

\end{lstlisting}

\end{document}